\documentclass[11pt]{article}

\usepackage[preprint]{acl}

\usepackage{times}
\usepackage{latexsym}

\usepackage{amsmath}
\usepackage{booktabs}
\usepackage{enumitem}
\usepackage{tcolorbox}

\usepackage{amsmath}
\usepackage{amssymb}
\usepackage{multirow}
\usepackage[T1]{fontenc}

\usepackage[utf8]{inputenc}

\usepackage{microtype}

\usepackage{inconsolata}

\usepackage{graphicx}

\title{Attention-guided Fine-tuning of Multimodal Large Language Models Improves Chain-of-Thought Reasoning}

\author{Sanchit Sinha, Guangzhi Xiong, Bohan Liu, Zhenghao He and Aidong Zhang \\
University of Virginia}
\begin{document}
\maketitle
\begin{abstract}
The effectiveness of Chain-of-Thought (CoT) prompting in Multimodal Large Language Models (MLLMs) remains uncertain: across several \textit{visual reasoning} benchmarks, CoT prompting often \emph{degrades} performance compared to direct prompting. In this paper, we provide a systematic analysis of CoT behavior in three modern MLLM families across model scales on datasets requiring step-wise visual evidence. Our analysis identifies two recurring failure modes: \textit{premature answer commitment} and \textit{limited direct visual-token access} during rationale generation. We further find that standard CoT-style Supervised Fine-Tuning (CoT-SFT) can mitigate these issues only partially, while often increasing reliance on textual priors and reducing counterfactual visual dependence. Motivated by these findings, we propose \emph{Attentive-CoT (Att-CoT)}, an attention-guided fine-tuning objective that encourages CoT trajectories to delay answer commitment while maintaining sustained visual-token access. Att-CoT can be plugged into any CoT-SFT training run without architectural changes. Experiments on three visual reasoning benchmarks across six MLLMs show that Att-CoT enhances CoT performance over standard fine-tuning. 
\end{abstract}

\section{Introduction}
Multimodal Large Language Models (MLLMs) have revolutionized the field of visual reasoning, powering diverse and complex problems, such as autonomous driving \cite{Tian2024DriveVLM,Sima2024ECCV}, image captioning \cite{Hua_2025_CVPR}, scene understanding \cite{shen2022shortest}, etc. A plethora of complex MLLM architecture families of various scales have been proposed, such as LLaVA \cite{liu2023llava}, Qwen-VL \cite{qwen2.5-VL}, InternVL \cite{wang2025internvl3}, etc. At their core, an MLLM consists of an image encoder, a pre-trained Large Language Model (LLM), and an alignment module that maps vision embeddings to the LLM's token space. As the most dominant part of the MLLM architecture is the LLM, MLLMs inherit many of the intrinsic \textit{reasoning mechanisms} of LLMs, e.g., In-Context Learning (ICL) \cite{brown2020language}, Chain-of-Thought (CoT) reasoning \cite{wei2022emergent}, etc. Despite the potential, these mechanisms in MLLMs during \textit{multi-step visual reasoning} remain poorly understood, especially when compared to the growing body of analysis on text-only LLMs.

Chain-of-Thought (CoT), first proposed in \citet{Wei2022CoT} on LLMs and improved upon in \citet{wang2022self}, enables models to think step-by-step before answering a query, mimicking the human thought process. Multiple works have ascertained the effectiveness of CoT in LLMs \cite{Kojima2022ZeroShotReasoners,wang2022self}, making CoT a cornerstone of LLM reasoning research. However, unlike in the language-only setting, the effectiveness of CoT prompting in MLLMs is mixed \cite{turpin2023language}. It is observed that CoT prompting often fails to improve performance and can even \emph{degrade} it relative to eliciting a simple \emph{direct} answer that omits intermediate steps, an observation that \textbf{stands in contrast to LLMs}. Some recent work, such as \citet{sun2025mitigating, xu2025llava, zhang2025improve}, has noticed this issue for MLLMs; however, it has not systematically analyzed the phenomenon. Intuitively, the reasoning of MLLMs relies heavily on both \emph{visual tokens} and \emph{text tokens}, since correct reasoning traces must depend significantly on understanding the image, including its compositionality and spatial relations.

To improve and impart better CoT reasoning in MLLMs, recent research like \citet{chen2024measuring, xu2025llava} has focused on \textbf{Supervised Fine-Tuning} (SFT) on meticulously labeled reasoning traces in addition to the final answer. In this setting, the model is trained to predict the correct reasoning trace in addition to the output, imitating step-by-step rationales collected from humans or distilled from stronger teacher models (e.g. GPT-4o). Such fine-tuning procedures (denoted as CoT-SFT\footnote{Sometimes also referred to as distillation. To avoid confusion with `model distillation', we omit this term and utilize CoT-SFT.}) have been widely adopted in recent MLLMs \cite{zhang2025improve, zhang2023multimodal, chen2024measuring} to boost performance on complex visual reasoning benchmarks, and are often treated as a default recipe for improving MLLM reasoning. Although some works claim that it improves reasoning in MLLMs \cite{zhang2025improve}, empirical results are mixed. CoT-SFT often improves formatting and step-by-step fluency, but it does not guarantee CoT's reliance on visual evidence in reasoning. In practice, as state-of-the-art systems are often deployed with either off-the-shelf or SFT models, they remain largely unexplored, requiring a systematic study.

In this paper, we diagnose the shortcomings of CoT in MLLMs using three complementary perspectives. First, we evaluate the performance gap between direct and CoT prompting on visually intensive benchmarks. Second, we track token-level answer probability as CoT tokens are generated and quantify it through \emph{Early Answer Commitment}, i.e., how early the model assigns high probability to the final correct answer during rationale generation. Third, we measure visual evidence usage through two views: a counterfactual \emph{Visual Dependence} probe, which replaces the image while keeping the textual context fixed, and a visual-token attention analysis, which diagnoses whether reasoning tokens directly access image-token representations during reasoning. We treat visual-token attention values as a measurable proxy for direct access to visual representations during CoT reasoning. While attention alone does not establish faithfulness \cite{jain2019attention}, we validate its utility by pairing it with counterfactual visual-dependence probes and downstream accuracy/OOD (out of distribution) evaluations. Together, these analyses reveal a consistent failure mode: models often commit to the final answer early, while intermediate CoT tokens show weak visual dependence and limited direct visual-token access compared to answer emission, promoting worse CoT performance over direct decoding.

Motivated by these observations, we propose a more principled view of CoT-SFT for MLLMs. We argue that a good reasoning MLLM should exhibit two desirable properties: (i) \emph{delayed answer commitment}, where answer probability should increase gradually with CoT token generation, and (ii) \emph{sustained visual access}, where intermediate reasoning tokens should remain conditioned on image evidence rather than deferring visual processing to the final answer span. Guided by these desiderata, we propose \textbf{Attentive-CoT}, a simple fine-tuning objective that explicitly discourages early answer commitment while encouraging visual-token access during rationale generation. We then validate whether this intervention improves reasoning not only through attention-based diagnostics, but also through downstream accuracy, Direct-CoT gap, OOD performance, and counterfactual visual dependence. More concretely, our contributions are:

\begin{itemize}[parsep=0pt, itemsep=0pt, topsep=0pt]
\item We systematically analyze the shortcomings of Chain-of-Thought (CoT) performance through probabilistic, counterfactual, and visual-token-access perspectives in MLLMs. We attribute these failures in part to \emph{premature answer commitment} and \emph{weak visual evidence usage} during rationale generation.

\item We introduce \textbf{Attentive-CoT}, a simple, model-agnostic fine-tuning objective that improves CoT-SFT by encouraging \emph{delayed answer commitment} and \emph{sustained visual-token access} over reasoning tokens.

\item Across three MLLM families of two distinct scales and three visual reasoning datasets, Attentive-CoT improves CoT accuracy over the standard CoT-SFT objective and yields CoT traces that are less prone to early commitment and more dependent on image evidence under counterfactual visual interventions.
\end{itemize}

\section{Related Work}
Chain-of-thought (CoT) prompting elicits multi-step reasoning in LLMs \cite{Wei2022CoT,wang2022self}, but in MLLMs, it can even degrade reasoning quality. This motivates evaluation beyond accuracy, for example \cite{lobo2025impact,chen2024measuring} propose metrics and interventions to assess and improve CoT behavior. SFT on intermediate traces alongside answers \cite{ren2024learning,shuttleworth2024lora,sinha2025chart} has been proposed. Beyond generic rationale SFT, recent work targets improved multimodal CoT by strengthening visual grounding, e.g., mitigating \emph{visual forgetting} via carried-forward visual conditioning \cite{sun2025mitigating} or explicitly training step-by-step reasoning \cite{xu2024llava} - choices adapted from LLM fine-tuning approaches. Faithfulness analyses argue that slow thinking should maintain evidence alignment throughout the chain \cite{chen2025bring,uppaal2025journey,sinha2025coco}. Our work analyzes CoT-SFT in MLLMs and introduces a simple, architecture-agnostic objective to improve CoT behavior. To the best of our knowledge, our work is the first to \emph{improve upon the CoT-SFT objective in MLLMs}, and no widely adopted SFT objective is explicitly designed for improving CoT behavior.

\noindent\textbf{Comparisons with Related Work.}
While LLaVA-CoT~\cite{xu2025llava} and LLaVA-Reasoner-DPO~\cite{zhang2025improve} also fine-tune MLLMs to improve CoT performance, our goal and training signal are fundamentally different. Both works solely optimize task performance using a standard SFT optimization objective as the first step. Further, LLaVA-CoT utilizes a test-time response optimization method to improve performance during inference, while LLaVA-Reasoner-DPO utilizes a DPO fine-tuning step (with synthetic data) after the SFT phase. Unlike them, we propose improvement over the SFT objective itself using probabilistic and mechanistic constraints drawn from CoT diagnostics. This makes our approach explicitly \emph{mechanism-targeted} - instead of \emph{what} rationale to generate, we supervise \emph{how} reasoning unfolds. As both approaches are trained primarily on earlier generations of the LLaVA family of models with limited out-of-the-box CoT performance (e.g. LLaVA 90B model underperforms Qwen2VL-7B on CoT despite containing 12 times more parameters - refer Figure-1 in \cite{xu2025llava}) , we replicate the method on current state-of-the-art MLLMs, which are pre-trained to be CoT-native off-the-shelf. The detailed replication setup can be found in the Appendix.

\section{Quantifying CoT Reasoning Dynamics}
\label{sec:analysis}

\subsection{Performance: Direct vs CoT}
\noindent\textbf{Motivation.}
We first evaluate downstream performance on three datasets: ChartQA \cite{masry2022chartqa}, CLEVR \cite{Johnson_2017_CVPR} and CV-Bench \cite{tong2024cambrian}, following the vision-centric categorization in \citep{tong2024cambrian}. ChartQA requires extracting numerical values from charts, CLEVR emphasizes compositional queries such as counts, spatial comparisons, logical combinations, etc., and CV-Bench tests fine-grained visual understanding. Table~\ref{tab:direct-vs-cot} reports the accuracy of \emph{direct} and \emph{CoT} prompting across Qwen2.5-VL, denoted as QVL, and InternVL3.5, denoted as IVL.

\begin{table}
    \caption{Direct vs CoT prediction performance.}
    \small
    \centering
    \resizebox{0.75\linewidth}{!}{
    \begin{tabular}{lcc|cc}
        \toprule
        \multirow{2}{*}{\bf Dataset} &
        \multicolumn{2}{c}{QVL-7B} &
        \multicolumn{2}{c}{IVL-4B} \\
        \cmidrule(lr){2-3} \cmidrule(lr){4-5}
        & \bf Direct & \bf CoT & \bf Direct & \bf CoT \\
        \midrule
        ChartQA  & \bf 83.44 & 79.23 & \bf 86.25 & 82.33 \\
        CLEVR    & \bf 86.25  & 85.71 & \bf 65.50  & 65.17 \\
        CV-Bench & \bf 70.75 & 66.96 & \bf 61.82 & 59.82  \\
        \bottomrule
    \end{tabular}
    }
    \label{tab:direct-vs-cot}
    \vspace{-10pt}
\end{table}

On all three visual-heavy datasets, CoT consistently \textbf{hurts} performance for both MLLMs. These results show that, unlike in text-only LLMs, \textit{CoT is not guaranteed to be superior} in MLLMs on visually grounded tasks. This motivates a closer analysis of how CoT changes reasoning dynamics and visual evidence usage in MLLMs.

\begin{figure}[h]
\centering
  \centering
  \includegraphics[width=\linewidth]{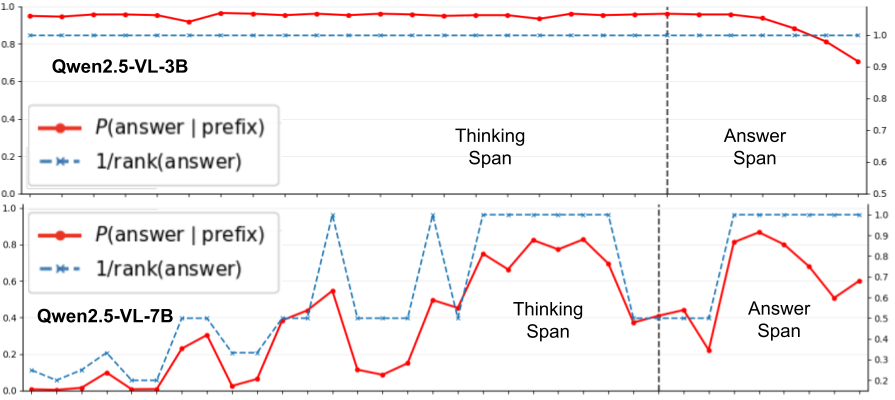}
\caption{Early answer commitment during CoT. We plot $\ell_t$ (denoted by `P(answer | prefix)') and the corresponding relative rank (i.e., rank in the ordered probability list) of the answer token over CoT timesteps for a ChartQA test sample. \textbf{Top:} QVL-3B is already highly confident before CoT generation starts. \textbf{Bottom:} QVL-7B remains less committed early and gradually increases confidence as the answer is computed. The vertical dashed line separates \texttt{think} and \texttt{answer} spans.}
\label{fig:base-answer-commit}
\vspace{-10pt}
\end{figure}

\subsection{Token-level Analysis}
To investigate the observed degradation, we probe how the model's certainty over the correct answer changes as CoT tokens are generated. Let each example be $(I,q,r^\star,y^\star)$, where $I$ is the image, $q$ is the question, $r^\star=(r^\star_1,\dots,r^\star_T)$ is a CoT rationale, and $y^\star=(y^\star_1,\dots,y^\star_K)$ is the gold answer. At step $t$, we construct $x_t = \bigl[I, q, r^\star_{1:t}, p\bigr]$, where $p$ is the probe prompt ``Therefore the answer is ''. For parameters $\theta$, we define probability assigned to the gold answer, normalized by answer length, as:
\begin{equation}
    \ell_t(x_t,y^\star)
    =
    \frac{1}{K}\sum_{k=1}^{K}
    p_\theta(y^\star_k \mid x_t,y^\star_{<k}).
    \label{eq:cot-probe-ll}
\end{equation}
Intuitively, $\ell_t$ measures how confident the model is in the final answer if queried with the current CoT prefix. During generation, we enforce an explicit \texttt{<think>} and \texttt{<answer>} output structure. Figure~\ref{fig:base-answer-commit} shows $\ell_t$ across CoT tokens for a ChartQA sample. Smaller MLLMs are highly confident in the correct answer very early, even before reasoning through the CoT tokens. In contrast, larger MLLMs remain relatively less committed early and gradually increase answer probability, peaking near answer emission. This suggests that MLLMs often generate rationales around an already-selected answer.

\noindent\textbf{Metric: Early Commitment AUC (EC-AUC).}
\label{sec:token-probes}
To quantify premature answer commitment, for each step $t\in\{0,\dots,T\}$, with $x_0=[I,q,p]$:
\begin{equation}\small
\label{eq:ac-auc}
    \mathrm{EC\text{-}AUC}
    =
    \frac{1}{T}
    \sum_{t=0}^{T}
    \left(
    \log \ell_t(x_t,y^\star)
    -
    \log \ell_t(x_t,\text{\texttt{[<pad>]}})
    \right).
\end{equation}
The \texttt{<pad>} baseline follows \citep{kim2020interpretation}; its probability is relatively stable across CoT generation and reduces sensitivity to distribution shifts. We compute EC-AUC under two rationale sources: \emph{free-form CoT}, sampled from the model itself, and \emph{ground-truth CoT}, using annotated rationales (Refer to Appendix for details). Lower EC-AUC indicates that answer confidence rises gradually over CoT rather than appearing prematurely. Table~\ref{tab:cot-analysis} reports EC-AUC across datasets, model families, and scales. Smaller MLLMs generally show higher EC-AUC, suggesting that they often commit to the answer early. Larger models show lower EC-AUC and behavior more consistent with gradual reasoning that also correlates with better performance.

\begin{table}[h]
\centering
\caption{CoT analysis metrics under CoT prompting for six off-the-shelf MLLMs. Lower EC-AUC indicates less premature commitment; lower ATR indicates more visual-token access during reasoning relative to answer emission; higher Vis-Dep indicates stronger counterfactual image dependence.}
\setlength{\tabcolsep}{4pt}
\resizebox{0.9\linewidth}{!}{%
\begin{tabular}{ll|c|cc|c|c}
\toprule
\multirow{2}{*}{\textbf{Dataset}} & \multirow{2}{*}{\textbf{Model}}
& \multirow{2}{*}{\textbf{Acc.} $\uparrow$}
& \multicolumn{2}{c|}{\textbf{EC-AUC} $\downarrow$}
& \multirow{2}{*}{\textbf{ATR} $\downarrow$}
& \multirow{2}{*}{\textbf{Vis-Dep} $\uparrow$} \\
\cmidrule(lr){4-5}
& & & Freeform & Gold & & \\
\midrule
\multirow{6}{*}{\textbf{ChartQA}}
  & QVL-3B    & 78.44 & 0.939 & 0.958 & 0.875 & 0.06 \\
  & IVL-4B    & 82.23 & 0.893 & 0.937 & 0.709 & 0.72 \\
  & Gemma-4B  & 74.56 & 0.858 & 0.892 & 0.714 & 0.69 \\
  & QVL-7B    & 79.23 & 0.765 & 0.743 & 0.793 & \textbf{0.97} \\
  & IVL-8B    & \textbf{83.43} & \textbf{0.238} & 0.602 & \textbf{0.661} & 0.76 \\
  & Gemma-12B & 78.45 & 0.412 & \textbf{0.592} & 0.674 & 0.84 \\
\midrule
\multirow{6}{*}{\textbf{CLEVR}}
  & QVL-3B    & 80.17 & 0.856 & 0.904 & 0.547 & 0.49 \\
  & IVL-4B    & 65.17 & 0.637 & 0.714 & 0.528 & 0.54 \\
  & Gemma-4B  & 72.92 & 0.748 & 0.812 & 0.642 & 0.58 \\
  & QVL-7B    & 85.71 & 0.772 & 0.766 & 0.580 & 0.52 \\
  & IVL-8B    & \textbf{86.67} & \textbf{0.224} & \textbf{0.473} & \textbf{0.513} & \textbf{0.59} \\
  & Gemma-12B & 85.24 & 0.483 & 0.565 & 0.534 & 0.58 \\
\midrule
\multirow{6}{*}{\textbf{CV-Bench}}
  & QVL-3B    & 60.71 & 0.976 & 0.978 & 0.413 & 0.21 \\
  & IVL-4B    & 59.82 & 0.971 & 0.978 & 0.494 & 0.18 \\
  & Gemma-4B  & 62.55 & 0.885 & 0.924 & 0.412 & 0.48 \\
  & QVL-7B    & 66.96 & 0.932 & 0.863 & 0.296 & 0.66 \\
  & IVL-8B    & 73.25 & 0.558 & 0.503 & \textbf{0.231} & 0.47 \\
  & Gemma-12B & \textbf{74.85} & \textbf{0.527} & \textbf{0.482} & 0.252 & \textbf{0.71} \\
\bottomrule
\end{tabular}%
}
\vspace{-5pt}
\label{tab:cot-analysis}
\end{table}

\begin{figure}[h]
    \centering
    \includegraphics[width=0.8\linewidth]{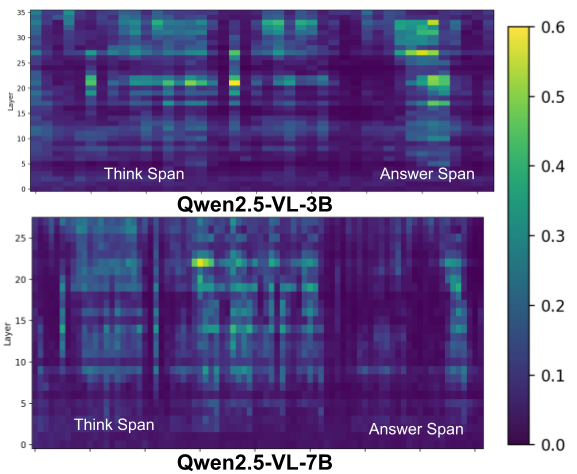}
    \caption{VTA values plotted across generation steps for each layer of the MLLM. \textbf{Left:} QVL-3B shows sparse VTA during CoT generation and higher values near the answer span. \textbf{Right:} QVL-7B shows more balanced VTA across CoT generation and answer emission.}
    \label{fig:attention-maps}
    \vspace{-10pt}
\end{figure}

\noindent\textbf{Metric: Visual Dependence Score.}
EC-AUC measures answer confidence, but not whether the CoT process depends on the image. We intervene by replacing the image with a zero image as $
x_t = [I,q,r^\star_{1:t}],  
x^{\text{zero}}_t = [\tilde I,q,r^\star_{1:t}],
$
where $\tilde I$ is zero image of same size. We define visual dependence (Vis-Dep) as:
\begin{equation}
    \frac{1}{T}\sum_{t=1}^{T}
    \left(
    \log \ell_t(x_t,y^\star)
    -
    \log \ell_t(x_t^{\text{zero}},y^\star)
    \right).
\end{equation}
Higher Vis-Dep indicates that replacing the image substantially changes the model's answer likelihood, implying stronger dependence on visual evidence. Table~\ref{tab:cot-analysis} reports Vis-Dep over the \texttt{think} span. Smaller models consistently show lower Vis-Dep than larger models, suggesting that their CoT generation is less image-dependent and more driven by textual priors. Combined with EC-AUC, this reinforces that smaller MLLMs often generate rationales weakly conditioned on the image. Results with alternative baselines are in the Appendix.

\subsection{Visual Token Attention Analysis}
\label{sec:vis-attn}
The token-generation experiments above quantify how answer confidence evolves and whether predictions depend on the image, but they do not reveal whether generated reasoning tokens directly access visual representations. We therefore analyze self-attention over visual tokens during CoT generation as a diagnostic of direct visual-token access. Let an MLLM have $L$ decoder layers and $H$ attention heads. For generated token position $t$, visual-token set $\mathcal{V}$, and attention weight $A^{(l,h)}_{t,j}$, we define:
\begin{equation}
    \text{VTA}^{l}(t)
    =
    \frac{1}{H}
    \sum_{h=1}^{H}
    \sum_{j\in\mathcal{V}}
    A^{(l,h)}_{t,j}.
    \label{eq:vta}
\end{equation}

Figure~\ref{fig:attention-maps} visualizes VTA across generation steps and layers. Smaller models show intermittent visual attention during the \texttt{think} span, followed by sharp peaks in the \texttt{answer} span. This suggests that they often defer direct visual-token access until answer emission, rather than maintaining it throughout the reasoning trajectory. Larger models are more balanced, which aligns with early-commitment behavior: models that commit early generate CoT tokens around the answer rather than progressively conditioning on visual evidence.

\noindent\textbf{Metric: Answer-Think Ratio (ATR).}
To quantify this imbalance, we aggregate VTA over the \texttt{think} and \texttt{answer} spans:
\begin{equation}
    \text{ATR}
    =
    \frac{
    \sum_{t\in\texttt{answer}}\sum_{l=1}^{L}\text{VTA}^{l}(t)
    }{
    \sum_{t\in\texttt{think}}\sum_{l=1}^{L}\text{VTA}^{l}(t)
    }.
\end{equation}
A lower ATR indicates that the model allocates greater visual-token attention during reasoning than during answer emission. Table~\ref{tab:cot-analysis} shows that models with lower EC-AUC tend to have higher VTA during the \texttt{think} span and lower ATR, indicating a correlation between delayed answer commitment and sustained visual-token access. Conversely, models whose visual attention peaks mainly in the \texttt{answer} span tend to exhibit weaker CoT performance.

Overall, our analyses indicate that CoT failures in MLLMs stem from a structural mismatch between reasoning dynamics, visual evidence usage, and answer commitment. Based on these findings, we identify two desiderata for CoT-capable MLLMs: \textbf{(1) Delayed Answer Commitment}, where answer confidence should increase progressively over the reasoning trajectory; and \textbf{(2) Sustained Visual Attention}, where intermediate reasoning tokens should remain consistently conditioned on image evidence.
\section{Proposed Method: Attentive-CoT}
\label{sec:method}
\vspace{-5pt}
Our analysis in Section~\ref{sec:analysis} shows that, for many MLLMs and datasets, CoT tokens are only weakly grounded in the image and often exhibit premature answer commitment. Motivated by the desiderata, we introduce \textbf{Attentive-CoT (Att-CoT)}, a substantial improvement over the SFT objective that (i) discourages early answer commitment during rationale generation and (ii) encourages sustained visual reliance while generating both reasoning and answer tokens. Attentive-CoT is architecture-agnostic and can be applied on top of any MLLM.

We assume a dataset of CoT-annotated training examples
$\mathcal{D} = \{(I, q, r^\star, y^\star)\}$,
where $I$ is the input image, $q$ is the question, $r^\star = (r^\star_1,\dots,r^\star_T)$ is the gold CoT rationale, and $y^\star = (y^\star_1,\dots,y^\star_K)$ is the gold answer sequence.
Let $p_\theta(\cdot \mid \cdot)$ denote the model’s next-token distribution for parameters $\theta$.
The standard CoT-style SFT loss $\mathcal{L}_{\text{SFT}}(\theta)$ is:
\begin{equation*}
\begin{aligned}
    \mathbb{E}_{(I,q,r^\star,y^\star) \sim \mathcal{D}} 
     \Bigg[
         -\sum_{t=1}^{T}
            \log p_\theta\!\left(r^\star_t \mid I, q, r^\star_{<t}\right) 
            \\
            - \sum_{k=1}^{K}
            \log p_\theta\!\left(y^\star_k \mid I, q, r^\star, y^\star_{<k}\right)
      \Bigg].
\end{aligned}
\label{eq:sft-loss}
\end{equation*}

\noindent \textbf{Attentive-CoT} augments $\mathcal{L}_{\text{SFT}}$ with two auxiliary terms derived directly from our desiderata: \textbf{delayed answer commitment} and \textbf{sustained visual reliance during reasoning}. For a training sample $(I,q,r^\star,y^\star)$, let $s = (s_1,\dots,s_{T+K})$ denote the concatenation of CoT and answer tokens, i.e., $s_{1:T}=r^\star$ and $s_{T+1:T+K}=y^\star$. Let $\mathcal{R}=\{1,\dots,T\}$ index CoT positions and let $t_{\text{ans}} = T+1$ denote the position of the first answer token $y^\star_1$. As in Section~\ref{sec:vis-attn}, for each position $t$, we compute the average attention mass assigned to visual tokens, denoted $A_{t,\text{vis}}\in[0,1]$, by averaging over layers/heads and summing attention weights ($A$) to visual-token indices $\mathcal{V}$ as:
\begin{equation}
\label{eq:avis}
A_{t,\mathrm{vis}}
\;=\;
\frac{1}{L\cdot H}\sum_{l=1}^{L}\sum_{h=1}^{H}\sum_{j\in\mathcal{V}} A^{(l,h)}_{t,j}.
\end{equation}

\noindent\textbf{Delayed Answer Commitment.}
\label{sec:dc-loss}
To discourage premature answer commitment during rationale generation, we penalize cases where the model assigns high probability to the answer too early in the CoT. Specifically, for each rationale position $t\in \mathcal{R}$, we compute the probability $\pi_t$ of the answer tokens at each step:
\begin{equation}
\pi_t \;=\; p_\theta\!\left(y^\star_1 \mid I,q, s_{<t}\right),~~
t\in \mathcal{R}.
\label{eq:pi-def}
\end{equation}
We want the model to assign substantially higher probability to $y^\star_1$ only at the answer start $t_{\text{ans}}$ as:
\begin{equation}
\begin{aligned}
\mathcal{L}_{\text{DC}}(\theta)
= \mathbb{E}_{(I,q,r^\star,y^\star)\sim\mathcal{D}}
\Bigg[
\frac{1}{|\mathcal{R}|}
\sum_{t\in\mathcal{R}}
\log\!\Big(
1 + {} & \\
\exp\!\big(
\log \pi_t - \log \pi_{t_{\text{ans}}}
\big)
\Big)
\Bigg].
\end{aligned}
\label{eq:dc-loss}
\end{equation}
To minimize computation, in practice, we enforce $t\leq T/2$. Intuitively, minimizing $\mathcal{L}_{\text{DC}}$ encourages $\pi_{t_{\text{ans}}} \gg \pi_t$, i.e., the model should avoid committing to the answer until it reaches end of CoT. More details can be found in the Appendix.

\noindent\textbf{Sustained Visual Access.}
\label{sec:vg-loss}
Next, to mitigate the non-reliance of CoT tokens on the image (as quantified by Vis-Dep), we encourage the model to maximize attention to visual tokens while generating intermediate reasoning tokens.  Using the visual attention mass $A_{t,\text{vis}}$ at each CoT position:
\begin{equation}
\begin{aligned}
\mathcal{L}_{\text{VG}}(\theta)
&= \mathbb{E}_{(I,q,r^\star,y^\star)\sim\mathcal{D}}
\left[
-\frac{1}{|\mathcal{R}|}
\sum_{t\in \mathcal{R}}
\log\!\big(A_{t,\text{vis}}\big)
\right].
\end{aligned}
\label{eq:vg-loss}
\end{equation}
This term biases training toward allocating meaningful attention mass to visual tokens while producing the rationale, promoting grounding throughout the \texttt{think} span.

\noindent\textbf{Overall Attentive-CoT Objective.}
\label{sec:overall-attcot}
The final Attentive-CoT objective is:
\begin{equation}
\mathcal{L}_{\text{Att-CoT}}(\theta)
=
\mathcal{L}_{\text{SFT}}(\theta)
+ \,\mathcal{L}_{\text{VG}}(\theta)
+ \mathcal{L}_{\text{DC}}(\theta)
\label{eq:attcot-loss}
\end{equation}

\section{Experiments}
\label{sec:experiments}
\vspace{-4pt}
In this section, we evaluate Att-CoT along five facets of CoT behavior: (i) performance, (ii) early answer commitment, (iii) attention on vision tokens, (iv) visual dependence, and (v) the Direct-CoT gap. We group these into two categories - early answer commitment and visual-token attention are defined over the optimization terms, so we report them as verification that the objective takes effect, while accuracy, visual dependence, and the Direct–CoT gap constitute our independent evidence that Att-CoT improves reasoning. We report results on \textbf{ChartQA}, \textbf{CLEVR} and \textbf{CV-Bench} across six LVLM backbones (\textbf{QVL-3B/7B} \cite{qwen2.5-VL}, \textbf{IVL-4B/8B} \cite{wang2025internvl3} and \textbf{Gemma-4B/12B} \cite{team2025gemma}) in Tables~\ref{tab:main-results-consolidated} and \ref{tab:main-results-wo-ood}. We compare our approach to LLaVA-CoT \cite{xu2025llava} and LLaVA-Reasoner-DPO (LLaVA-R) \cite{zhang2025improve}. Baseline replication details are detailed in the Appendix.

\begin{table*}[t]
\centering
\small
\setlength{\tabcolsep}{5pt}
\resizebox{0.87\textwidth}{!}{%
\begin{tabular}{ll|ccccc|ccccc}
\toprule
\multirow{2}{*}{\textbf{Model}} &
\multirow{2}{*}{\textbf{Method}} &
\multicolumn{5}{c|}{\textbf{ChartQA / EvoCharts}} &
\multicolumn{5}{c}{\textbf{CLEVR / Super-CLEVR}} \\
\cmidrule(lr){3-7} \cmidrule(lr){8-12}
& &
\textbf{Acc.}$\uparrow$ &
\textbf{OOD}$\uparrow$ &
\textbf{EC-F}$\downarrow$ &
\textbf{Vis-Dep}$\uparrow$ &
\textbf{$\Delta_{\mathrm{CoT}}$}$\downarrow$ &
\textbf{Acc.}$\uparrow$ &
\textbf{OOD}$\uparrow$ &
\textbf{EC-F}$\downarrow$ &
\textbf{Vis-Dep}$\uparrow$ &
\textbf{$\Delta_{\mathrm{CoT}}$}$\downarrow$ \\
\midrule

\multirow{4}{*}{QVL-3B}
& SFT
& 81.25 & 45.53 & 0.925 & 0.62 & 4.32
& 84.20 & 54.80 & 0.897 & 0.41 & 1.45 \\
& LLaVA-CoT
& 78.33 & 42.84 & 0.892 & 0.55 & --
& 80.20 & 52.90 & 0.876 & 0.38 & -- \\
& LLaVA-R
& 80.15 & 46.11 & 0.784 & 0.63 & --
& 82.80 & 54.50 & 0.812 & 0.40 & -- \\
& \textbf{Att-CoT}
& \textbf{83.45} & \textbf{50.24} & \textbf{0.642} & \textbf{0.71} & \textbf{2.14}
& \textbf{86.50} & \textbf{57.10} & \textbf{0.643} & \textbf{0.43} & \textbf{-0.58} \\

\midrule

\multirow{4}{*}{IVL-4B}
& SFT
& 86.39 & 57.69 & 0.925 & 0.59 & 3.85
& 65.50 & 52.60 & 0.925 & 0.65 & 8.86 \\
& LLaVA-CoT
& 84.81 & 55.21 & 0.843 & 0.44 & --
& 62.80 & 49.40 & 0.855 & 0.60 & -- \\
& LLaVA-R
& 85.64 & 57.22 & 0.762 & 0.58 & --
& 64.90 & 52.80 & 0.796 & 0.62 & -- \\
& \textbf{Att-CoT}
& \textbf{86.84} & \textbf{59.12} & \textbf{0.502} & \textbf{0.72} & \textbf{3.05}
& \textbf{74.50} & \textbf{54.70} & \textbf{0.754} & \textbf{0.71} & \textbf{-0.20} \\

\midrule

\multirow{4}{*}{Gemma-4B}
& SFT
& 80.44 & 44.92 & 0.894 & 0.67 & 3.64
& 79.80 & 55.10 & 0.882 & 0.45 & 5.59 \\
& LLaVA-CoT
& 78.62 & 40.84 & 0.846 & 0.59 & --
& 76.40 & 54.10 & 0.819 & 0.36 & -- \\
& LLaVA-R
& 79.35 & 45.68 & 0.855 & 0.64 & --
& 78.70 & 55.20 & 0.842 & 0.40 & -- \\
& \textbf{Att-CoT}
& \textbf{82.54} & \textbf{48.22} & \textbf{0.624} & \textbf{0.78} & \textbf{1.04}
& \textbf{83.10} & \textbf{56.50} & \textbf{0.806} & \textbf{0.55} & \textbf{1.82} \\

\midrule

\multirow{4}{*}{QVL-7B}
& SFT
& 86.61 & 56.25 & 0.572 & 0.60 & 2.05
& 85.50 & 60.40 & 0.478 & 0.35 & 0.75 \\
& LLaVA-CoT
& 80.35 & 55.21 & 0.561 & 0.48 & --
& 83.20 & 58.90 & 0.412 & 0.31 & -- \\
& LLaVA-R
& 84.81 & 57.22 & 0.476 & 0.57 & --
& 83.90 & 61.10 & 0.360 & 0.35 & -- \\
& \textbf{Att-CoT}
& \textbf{86.82} & \textbf{58.55} & \textbf{0.315} & \textbf{0.83} & \textbf{2.15}
& \textbf{87.40} & \textbf{63.20} & \textbf{0.231} & \textbf{0.52} & \textbf{-1.22} \\

\midrule

\multirow{4}{*}{IVL-8B}
& SFT
& \textbf{91.07} & 60.57 & 0.415 & 0.51 & \bf 2.78
& 82.10 & 54.20 & 0.515 & -0.03 & 0.33 \\
& LLaVA-CoT
& 85.66 & 55.21 & 0.407 & 0.43 & --
& 80.60 & 53.10 & 0.527 & -0.01 & -- \\
& LLaVA-R
& 88.24 & 57.48 & 0.324 & 0.50 & --
& 82.00 & 54.60 & 0.464 & 0.08 & -- \\
& \textbf{Att-CoT}
& 90.85 & \textbf{61.72} & \textbf{0.192} & \textbf{0.69} & 2.66
& \textbf{83.70} & \textbf{56.90} & \textbf{0.320} & \textbf{0.18} & \textbf{-1.24} \\

\midrule

\multirow{4}{*}{Gemma-12B}
& SFT
& 89.45 & 55.21 & 0.482 & \textbf{0.75} & 1.63
& 81.80 & 60.60 & 0.502 & 0.11 & 3.58 \\
& LLaVA-CoT
& 87.24 & 52.46 & 0.444 & 0.63 & --
& 80.60 & 59.30 & 0.516 & 0.09 & -- \\
& LLaVA-R
& 88.57 & 53.69 & 0.387 & 0.65 & --
& 81.40 & 61.20 & 0.438 & 0.22 & -- \\
& \textbf{Att-CoT}
& \textbf{90.72} & \textbf{57.22} & \textbf{0.282} & 0.74 & \textbf{0.32}
& \textbf{86.30} & \textbf{62.90} & \textbf{0.347} & \textbf{0.32} & \textbf{-1.64} \\

\bottomrule
\end{tabular}%
}
\caption{
Consolidated comparison across ChartQA/EvoCharts and CLEVR/Super-CLEVR. 
Rows are grouped by model, with each method reported once and dataset-specific metrics shown in left-right blocks. 
EC-F denotes free-form EC-AUC. Gold-rationale EC-AUC and ATR are visualized separately in Figure~\ref{fig:metric-lines}. All results averaged over 3 seeds with std dev. $\leq$0.02.
}
\label{tab:main-results-consolidated}
\vspace{-10pt}
\end{table*}

\subsection{Task Performance}
\label{sec:exp-performance}
We report the performance of baselines and Att-CoT on the test set from each training data set (Refer Appendix).  Additionally, we report performance on OOD datasets for chart reasoning EvoCharts \cite{huang2025evochart} and for spatial reasoning task - Super-CLEVR \cite{li2023super} marked as \textbf{OOD} in Tables~\ref{tab:main-results-consolidated}. Att-CoT improves accuracy over SFT across all model families on test sets as shown in `Accuracy' column (Tables~\ref{tab:main-results-consolidated} and \ref{tab:main-results-wo-ood}). We observe consistent gains on ChartQA of about 2-3\% over SFT and other baselines, while on OOD EvoCharts of about 2-5\%. Similarly, on CLEVR and Super-CLEVR, the improvements are substantial - 3-9\% and 3-4\%, respectively. Finally, on CV-Bench, the performance improvements are 2-4\% as well. Notably, Att-CoT outperforms LLaVA-CoT and LLaVA-R baselines substantially, suggesting a fundamental improvement over standard CoT-SFT setups. Performance on difficulty-aware splits can be found in the Appendix.

\subsection{Answer Commitment: EC-AUC}
\label{sec:exp-ecauc}
\begin{figure}[h]
    \centering
    \includegraphics[width=\linewidth]{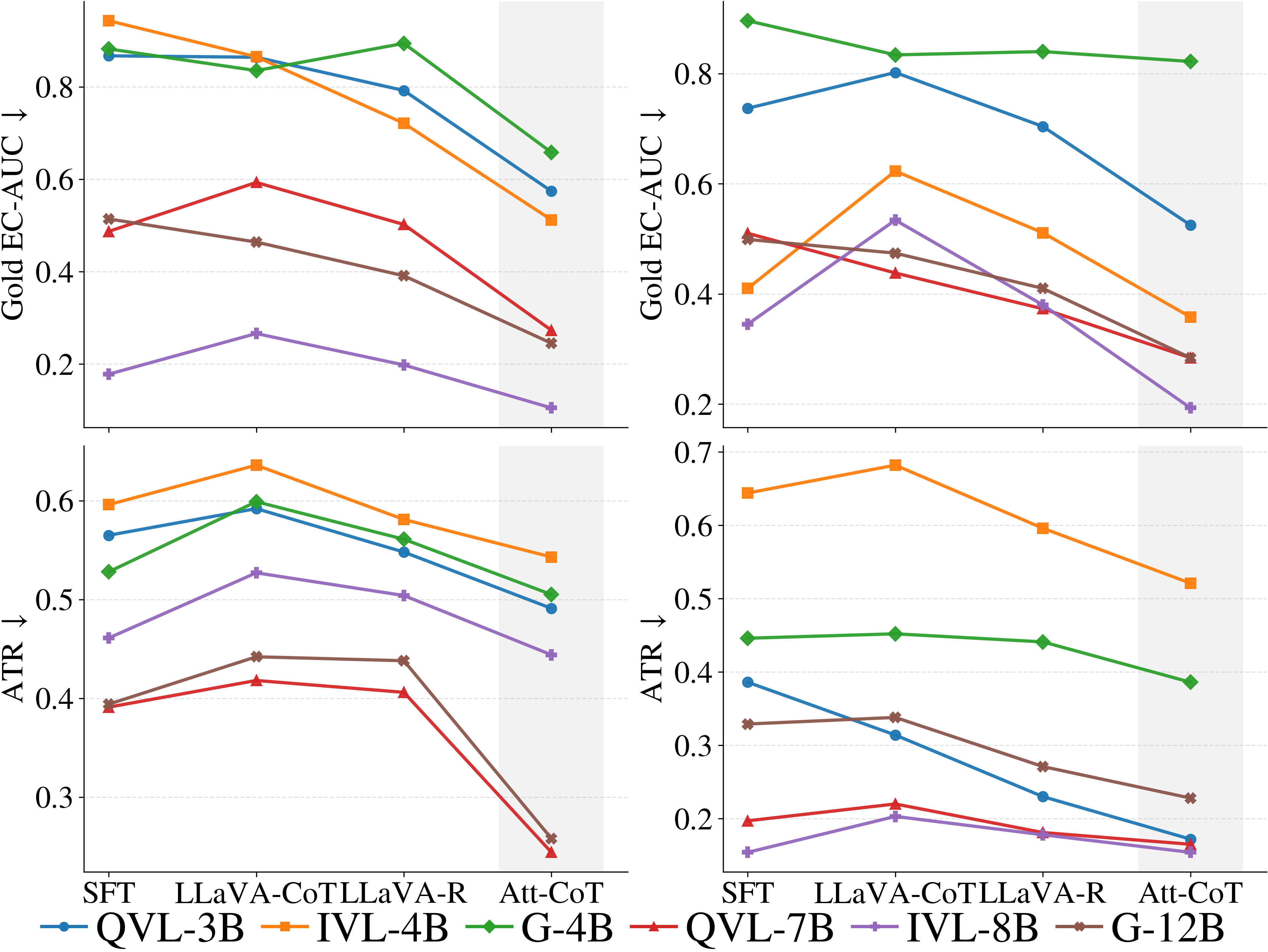}
    \caption{Gold EC-AUC and ATR across baselines. Left: ChartQA, Right: CLEVR.}
    \label{fig:metric-lines}
    \vspace{-10pt}
\end{figure}

Next, we test whether Att-CoT achieves the first desideratum: \textbf{delayed answer commitment}. 
Across datasets and backbones, Att-CoT substantially reduces EC-AUC (Note gold EC-AUC is shown in Figure~\ref{fig:metric-lines}), indicating fewer premature answer spikes during CoT generation (Tables~\ref{tab:main-results-consolidated}, \ref{tab:main-results-wo-ood}). 
The reduction in EC-AUC values holds on ChartQA, CLEVR, and CV-Bench, demonstrating that Att-CoT consistently mitigates early commitment across domains and model families. Notably, the baselines LLaVA-CoT and LLaVA-R demonstrate mixed results, with EC-AUC dropping mildly in the majority of cases.

\subsection{Visual Attention Analysis}
\label{sec:exp-visattn}
Next, we test the second desideratum: \textbf{sustained visual reliance during reasoning}. 
Figure~\ref{fig:metric-lines} reports the Answer-Think Ratio (ATR) across model families and datasets. Lower values of ATR are correlated with higher performance on both in-domain and OOD datasets. Overall, these results indicate that Att-CoT increases \emph{visual attention allocation} during reasoning and reduces late-stage attention spikes confined to answer emission. We treat this as mechanistic evidence that the model accesses visual tokens throughout multi-step generation.

\subsection{Visual Dependence Score}
\label{sec:exp-visdep}
Our primary evidence for visual reliance is counterfactual visual dependence, which intervenes on the visual modality and measures changes in answer likelihood. Tables~\ref{tab:main-results-consolidated} and \ref{tab:main-results-wo-ood} show that standard SFT often amplifies reliance on textual priors, reducing visual dependence relative to the base models. In contrast, Att-CoT consistently increases Vis-Dep across datasets and backbones (Tables~\ref{tab:main-results-consolidated}--\ref{tab:main-results-wo-ood}), indicating stronger image reliance during multi-step reasoning. The gains support that Att-CoT improves reasoning-time visual reliance (Table~\ref{tab:main-results-consolidated}).

\begin{figure*}[t]
    \centering
    \includegraphics[width=0.65\textwidth]{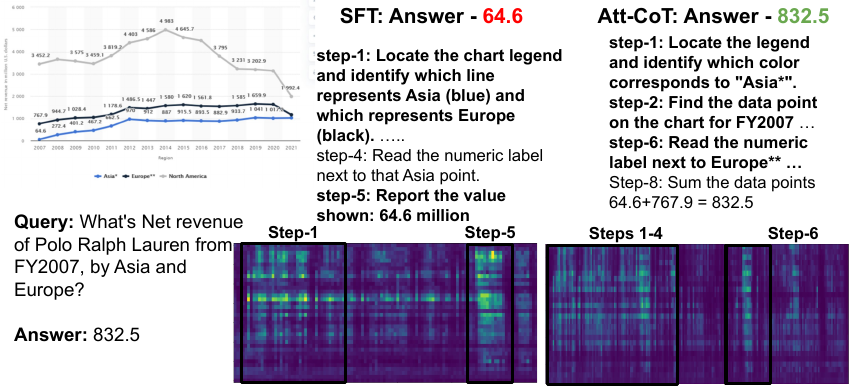}
    \caption{Qualitative comparison of SFT and Att-CoT on an example from ChartQA.}
    \label{fig:qualitative-example-1}
    \vspace{-10pt}
\end{figure*}

\vspace{-5pt}

\subsection{Improved Visual Faithfulness}

\begin{figure}[h]
    \centering
    \includegraphics[width=0.75\linewidth]{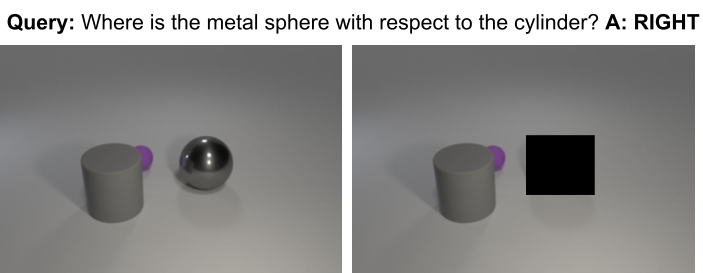}
    \caption{Example from the obfuscation dataset. The task-relevant visual region is masked.}
    \label{fig:eg-faithful}
    \vspace{-10pt}
\end{figure}

Next, we evaluate whether Att-CoT relies more strongly on task-relevant visual evidence, rather than only increasing visual-token attention. We construct a manually curated obfuscation set of 500 examples from CLEVR (350) and ChartQA (150), selecting examples that all compared methods answer correctly under the original image. For each example, we manually obfuscate the visual region required to answer the query, as shown in Figure~\ref{fig:eg-faithful}, while keeping the question unchanged.

We measure the drop in confidence assigned to the original correct answer after obfuscation. A larger confidence drop indicates that the model's original prediction was more dependent on the relevant visual evidence. As shown in Table~\ref{tab:example-faithful}, Att-CoT exhibits substantially larger confidence drops than SFT, LLaVA-CoT, and LLaVA-R on both CLEVR and ChartQA. This provides additional causal evidence that Att-CoT improves visual reliance.

\begin{table}[h]
\centering
\caption{Faithfulness under task-relevant visual obfuscation. We report confidence drop on the correct answer token. Higher is better.}
\label{tab:example-faithful}
\resizebox{0.75\linewidth}{!}{\begin{tabular}{c|cccc}
\toprule
\textbf{Dataset} & \textbf{SFT} & \textbf{L-CoT} & \textbf{L-R} & \textbf{Att-CoT} \\
\midrule
CLEVR   & 28.05 & 29.33 & 34.25 & \textbf{59.85} \\
ChartQA & 15.64 & 20.78 & 19.55 & \textbf{44.23} \\
\bottomrule
\end{tabular}}
\vspace{-15pt}
\end{table}

\subsection{Narrower Direct-CoT Performance Gap}
\label{sec:exp-directcot}
\vspace{-5pt}
Finally, we quantify whether Att-CoT narrows the gap between direct answering and CoT decoding. Let $\mathrm{Acc}_{\text{direct}}$ denote accuracy under direct prompting of a model SFT on final responses and $\mathrm{Acc}_{\text{cot}}$ denote accuracy under Att-CoT. We define the Direct-CoT gap as: $\Delta_{\text{CoT}}=\mathrm{Acc}_{\text{direct}} - \mathrm{Acc}_{\text{cot}}$. Lower values of $\Delta_{\text{CoT}}$ indicate a smaller disadvantage for CoT (and negative values indicate CoT outperforming direct).

\subsection{Qualitative Results}
Figure~\ref{fig:qualitative-example-1} shows SFT and Att-CoT in action on a sample from the ChartQA dataset. SFT concentrates visual attention on the initial identification step and the final reporting step, while exhibiting low visual attention during intermediate chart lookups and answering incorrectly. In contrast, Att-CoT maintains more uniform visual attention across the early lookup steps (Steps 1–4), supporting fine-grained value extraction and yielding the correct aggregated answer (Note: colormap is relative). We report more examples in the Appendix.


\begin{table}[t]
\centering
\caption{\textbf{Results on CV-Bench.} Att-CoT outperforms over all comparison baselines.}
\resizebox{0.44\textwidth}{!}{%
\begin{tabular}{l l|c|cc|c|c}
\toprule
\multirow{2}{*}{\textbf{Model}} & \multirow{2}{*}{\textbf{Method}}
& \multirow{2}{*}{\textbf{Accuracy} $\uparrow$}
& \multicolumn{2}{|c|}{\textbf{EC-AUC} $\downarrow$}
& \multirow{2}{*}{\textbf{ATR} $\downarrow$}
& \multirow{2}{*}{\textbf{Vis-Dep} $\uparrow$} \\
\cmidrule(lr){4-5}
& & & Freeform & Gold & & \\
\midrule
\multirow{4}{*}{QVL-3B} & SFT & 79.46 & 0.914 & 0.842 & 0.414 & 0.32 \\
& L-CoT  & 74.85 & 0.772  & 0.641 & 0.456 & 0.31 \\
& L-R  & 76.12 & 0.537  & 0.459 & 0.439 & 0.36 \\
& \textbf{Att-CoT}  & \bf 80.33   & \textbf{0.240} & \textbf{0.380} & \bf 0.402 & \bf 0.48 \\

\cmidrule(lr){1-7}
\multirow{4}{*}{IVL-4B} & SFT & 62.00 & 0.869 & 0.944 & 0.512 & -0.02 \\
& L-CoT  & 62.15 & 0.828  & 0.810 & 0.548 & 0.22 \\
& L-R  & 60.68 &  0.796 & 0.781 & 0.419 & 0.19 \\
& \textbf{Att-CoT}  & \textbf{63.44} & \bf 0.743  & \bf 0.725 & \bf 0.359 &  \bf 0.31 \\

\cmidrule(lr){1-7}
\multirow{4}{*}{G-4B} & SFT  & 66.85 & 0.599 & 0.587 & 0.455 & 0.45 \\
& L-CoT  & 64.09 & 0.571  & 0.555  & 0.476 & 0.41 \\
& L-R  & 67.28 & 0.497 & 0.470 & 0.405 & 0.51 \\
& \textbf{Att-CoT} &  \bf 74.24 & \bf 0.448  & \bf 0.416  & \bf 0.384 & \bf 0.56 \\

\cmidrule(lr){1-7}
\multirow{4}{*}{QVL-7B} & SFT  & 79.46 & 0.642 & 0.594 & 0.113 & 0.33 \\
& L-CoT  & 75.65 & 0.655  & 0.682 & 0.147 & 0.29 \\
& L-R  & 77.94 &  0.592 & 0.619 & 0.132 & 0.38 \\
& \textbf{Att-CoT} & \bf 81.25  & \bf 0.514 & \bf 0.382 & \bf 0.102 & \bf 0.58 \\

\cmidrule(lr){1-7}
\multirow{4}{*}{IVL-8B} & SFT  & 75.89 & 0.052 & 0.095 & 0.101 & 0.31 \\
& L-CoT  & 72.34 & 0.104  & 0.170 & 0.143 & 0.25 \\
& L-R  & 73.12 & 0.077  & 0.121 & 0.122 & 0.46 \\
& \textbf{Att-CoT} & \textbf{76.24} & \textbf{0.043} & \textbf{0.086} & \bf 0.100 & \bf 0.54 \\

\cmidrule(lr){1-7}
\multirow{4}{*}{G-12B} & SFT  & 75.49 & 0.270 & 0.295 & 0.176  & 0.47 \\
& L-CoT   & 73.55 &  0.363 & 0.389 & 0.221 & 0.44 \\
& L-R  & 76.05 & 0.246 & 0.291 & 0.173 & 0.47 \\
& \textbf{Att-CoT} & \bf 79.05  & \bf 0.186  & \bf 0.242 & \bf 0.135 & \bf 0.49 \\
\bottomrule
\end{tabular}%
}
\label{tab:main-results-wo-ood}
\vspace{-12pt}
\end{table}

\subsection{Ablation Study and Computation Requirements}
Finally, we conduct an ablation study to demonstrate the effect of $\mathcal{L}_\text{DC}$ and $\mathcal{L}_\text{VG}$. We focus on the QVL-3B model on the CLEVR dataset. $\mathcal{L}_{DC}$ substantially reduces early answer commitment and modestly improves visual reliance (Vis-Dep), while $\mathcal{L}_{VG}$ primarily increases visual dependence and shifts attention from the answer span to the think span (ATR), implying a balance between terms. 

\begin{table}[h]
    \centering
    \small
    \caption{Ablation Study on QVL-3B on CLEVR.}
    \resizebox{0.8\linewidth}{!}{
    \begin{tabular}{lcccc}
        \toprule
        \textbf{Setting} & \textbf{Accuracy} & \textbf{EC-AUC} & \textbf{ATR} & \textbf{Vis-Dep} \\
        \midrule
        SFT & 84.20 & 0.897 & 0.386 & 0.41 \\
        w/ $\mathcal{L}_{DC}$ & 84.70 & 0.686 & 0.314  & 0.42 \\
        w/ $\mathcal{L}_{VG}$ & 85.60 & 0.752 & 0.244 & \textbf{0.43} \\
        Att-CoT & \textbf{86.50} & \textbf{0.643} & \textbf{0.230} & \textbf{0.43} \\
        \bottomrule
    \end{tabular}}
    \label{tab:ablation}
    \vspace{-15pt}
\end{table} 

Att-CoT is asymptotically compute-matched to CoT-SFT and requires no additional model evaluations; implementation-level runtime and memory overhead depend on how attention statistics are exposed (depending on implementations). The added operation is a lightweight reduction over attention weights already produced during the forward pass, e.g., 
$\bar{A}=\frac{1}{LH}\sum_{\ell=1}^{L}\sum_{h=1}^{H} A^{(\ell,h)}$, where cost is negligible to dominant $O(LT^{2}d)$ compute. 

\section{Conclusion}
We revisit chain-of-thought (CoT) in MLLMs and find that, across multiple visual reasoning benchmarks, CoT prompting often reduces accuracy compared to direct decoding. Token-level probes and attention analyses show that CoT underutilizes image evidence: smaller models commit early, while larger models delay commitment but still attend weakly to visual tokens. To address this, we propose Attentive-CoT, an attention-guided fine-tuning objective that rewards stable, visually grounded CoT trajectories, consistently improving CoT accuracy. 

\section*{Limitations}
Att-CoT is designed as an improvement to the supervised CoT fine-tuning stage, and evaluating its interaction with RL-based post-training is left for future work. Our experiments use open-source MLLMs, and behavior may differ for larger closed-source systems with different pre-training and reasoning procedures. Finally, we limit our analysis to autoregressive MLLMs; our analysis might not apply directly to diverse architectures such as diffusion language models.


\bibliography{custom}

@article{chen2025bring,
  title={Bring reason to vision: Understanding perception and reasoning through model merging},
  author={Chen, Shiqi and Zhang, Jinghan and Zhu, Tongyao and Liu, Wei and Gao, Siyang and Xiong, Miao and Li, Manling and He, Junxian},
  journal={arXiv preprint arXiv:2505.05464},
  year={2025}
}

@article{sun2025mitigating,
    title = "Mitigating Visual Forgetting via Take-along Visual Conditioning for Multi-modal Long {C}o{T} Reasoning",
    author = "Sun, Hai-Long  and
      Sun, Zhun  and
      Peng, Houwen  and
      Ye, Han-Jia",
    editor = "Che, Wanxiang  and
      Nabende, Joyce  and
      Shutova, Ekaterina  and
      Pilehvar, Mohammad Taher",
    booktitle = "Proceedings of the 63rd Annual Meeting of the Association for Computational Linguistics (Volume 1: Long Papers)",
    month = jul,
    year = "2025",
    address = "Vienna, Austria",
    publisher = "Association for Computational Linguistics",
    url = "https://aclanthology.org/2025.acl-long.257/",
    doi = "10.18653/v1/2025.acl-long.257",
    pages = "5158--5171",
    ISBN = "979-8-89176-251-0"
}

@article{shuttleworth2024lora,
  title={Lora vs full fine-tuning: An illusion of equivalence},
  author={Shuttleworth, Reece and Andreas, Jacob and Torralba, Antonio and Sharma, Pratyusha},
  journal={arXiv preprint arXiv:2410.21228},
  year={2024}
}

@article{xu2024llava,
  title={Llava-o1: Let vision language models reason step-by-step},
  author={Xu, Guowei and Jin, Peng and Hao, Li and Song, Yibing and Sun, Lichao and Yuan, Li},
  journal={arXiv preprint arXiv:2411.10440},
  year={2024}
}

@article{shen2022shortest,
  title={Are shortest rationales the best explanations for human understanding?},
  author={Shen, Hua and Wu, Tongshuang and Guo, Wenbo and Huang, Ting-Hao'Kenneth'},
  journal={arXiv preprint arXiv:2203.08788},
  year={2022}
}

@inproceedings{huang2025evochart,
  title={Evochart: A benchmark and a self-training approach towards real-world chart understanding},
  author={Huang, Muye and Lai, Han and Zhang, Xinyu and Wu, Wenjun and Ma, Jie and Zhang, Lingling and Liu, Jun},
  booktitle={Proceedings of the AAAI Conference on Artificial Intelligence},
  volume={39},
  number={4},
  pages={3680--3688},
  year={2025}
}

@article{qwen2.5-VL,
  title={Qwen2.5-VL Technical Report},
  author={Bai, Shuai and Chen, Keqin and Liu, Xuejing and Wang, Jialin and Ge, Wenbin and Song, Sibo and Dang, Kai and Wang, Peng and Wang, Shijie and Tang, Jun and Zhong, Humen and Zhu, Yuanzhi and Yang, Mingkun and Li, Zhaohai and Wan, Jianqiang and Wang, Pengfei and Ding, Wei and Fu, Zheren and Xu, Yiheng and Ye, Jiabo and Zhang, Xi and Xie, Tianbao and Cheng, Zesen and Zhang, Hang and Yang, Zhibo and Xu, Haiyang and Lin, Junyang},
  journal={arXiv preprint arXiv:2502.13923},
  year={2025}
}

@inproceedings{masry2022chartqa,
  title={ChartQA: A Benchmark for Question Answering about Charts with Visual and Logical Reasoning},
  author={Masry, Ahmed and Do, Xuan Long and Tan, Jia Qing and Joty, Shafiq and Hoque, Enamul},
  booktitle={Findings of the Association for Computational Linguistics: ACL 2022},
  pages={2263--2279},
  year={2022}
}

@article{turpin2023language,
  title={Language models don't always say what they think: Unfaithful explanations in chain-of-thought prompting},
  author={Turpin, Miles and Michael, Julian and Perez, Ethan and Bowman, Samuel},
  journal={Advances in Neural Information Processing Systems},
  volume={36},
  pages={74952--74965},
  year={2023}
}

@article{team2025gemma,
  title={Gemma 3 technical report},
  author={Team, Gemma and Kamath, Aishwarya and Ferret, Johan and Pathak, Shreya and Vieillard, Nino and Merhej, Ramona and Perrin, Sarah and Matejovicova, Tatiana and Ram{\'e}, Alexandre and Rivi{\`e}re, Morgane and others},
  journal={arXiv preprint arXiv:2503.19786},
  year={2025}
}

@article{wang2025internvl3,
  title={Internvl3. 5: Advancing open-source multimodal models in versatility, reasoning, and efficiency},
  author={Wang, Weiyun and Gao, Zhangwei and Gu, Lixin and Pu, Hengjun and Cui, Long and Wei, Xingguang and Liu, Zhaoyang and Jing, Linglin and Ye, Shenglong and Shao, Jie and others},
  journal={arXiv preprint arXiv:2508.18265},
  year={2025}
}

@article{wei2022emergent,
  title={Emergent abilities of large language models},
  author={Wei, Jason and Tay, Yi and Bommasani, Rishi and Raffel, Colin and Zoph, Barret and Borgeaud, Sebastian and Yogatama, Dani and Bosma, Maarten and Zhou, Denny and Metzler, Donald and others},
  journal={arXiv preprint arXiv:2206.07682},
  year={2022}
}

@inproceedings{wang2022self,
title={Self-Consistency Improves Chain of Thought Reasoning in Language Models},
author={Xuezhi Wang and Jason Wei and Dale Schuurmans and Quoc V Le and Ed H. Chi and Sharan Narang and Aakanksha Chowdhery and Denny Zhou},
booktitle={The Eleventh International Conference on Learning Representations },
year={2023},
url={https://openreview.net/forum?id=1PL1NIMMrw}
}

@inproceedings{zhang2025improve,
  title={Improve vision language model chain-of-thought reasoning},
  author={Zhang, Ruohong and Zhang, Bowen and Li, Yanghao and Zhang, Haotian and Sun, Zhiqing and Gan, Zhe and Yang, Yinfei and Pang, Ruoming and Yang, Yiming},
  booktitle={Proceedings of the 63rd Annual Meeting of the Association for Computational Linguistics (Volume 1: Long Papers)},
  pages={1631--1662},
  year={2025}
}

@article{zhang2023multimodal,
  title={Multimodal chain-of-thought reasoning in language models},
  author={Zhang, Zhuosheng and Zhang, Aston and Li, Mu and Zhao, Hai and Karypis, George and Smola, Alex},
  journal={arXiv preprint arXiv:2302.00923},
  year={2023}
}

@inproceedings{chen2024measuring,
  title={Measuring and improving chain-of-thought reasoning in vision-language models},
  author={Chen, Yangyi and Sikka, Karan and Cogswell, Michael and Ji, Heng and Divakaran, Ajay},
  booktitle={Proceedings of the 2024 Conference of the North American Chapter of the Association for Computational Linguistics: Human Language Technologies (Volume 1: Long Papers)},
  pages={192--210},
  year={2024}
}

@article{tong2024cambrian,
  title={Cambrian-1: A fully open, vision-centric exploration of multimodal llms},
  author={Tong, Peter and Brown, Ellis and Wu, Penghao and Woo, Sanghyun and IYER, Adithya Jairam Vedagiri and Akula, Sai Charitha and Yang, Shusheng and Yang, Jihan and Middepogu, Manoj and Wang, Ziteng and others},
  journal={Advances in Neural Information Processing Systems},
  volume={37},
  pages={87310--87356},
  year={2024}
}

@inproceedings{lobo2025impact,
  title={On the impact of fine-tuning on chain-of-thought reasoning},
  author={Lobo, Elita and Agarwal, Chirag and Lakkaraju, Himabindu},
  booktitle={Proceedings of the 2025 Conference of the Nations of the Americas Chapter of the Association for Computational Linguistics: Human Language Technologies (Volume 1: Long Papers)},
  pages={11679--11698},
  year={2025}
}

@article{ren2024learning,
  title={Learning dynamics of llm finetuning},
  author={Ren, Yi and Sutherland, Danica J},
  journal={arXiv preprint arXiv:2407.10490},
  year={2024}
}

@article{uppaal2025journey,
  title={Journey Before Destination: On the importance of Visual Faithfulness in Slow Thinking},
  author={Uppaal, Rheeya and Htut, Phu Mon and Bai, Min and Pappas, Nikolaos and Qi, Zheng},
  journal={arXiv preprint arXiv:2512.12218},
  year={2025}
}

@article{liu2023llava,
  title={{LLaVA}: Large Language and Vision Assistant},
  author={Liu, Haotian and Li, Chunyuan and Wu, Qingyang and Li, Yong Jae},
  journal={arXiv preprint arXiv:2304.08485},
  year={2023}
}

@inproceedings{Tian2024DriveVLM,
  title        = {DriveVLM: The Convergence of Autonomous Driving and Large Vision-Language Models},
  author       = {Xiaoyu Tian and Junru Gu and Bailin Li and Yicheng Liu and Yang Wang and Zhiyong Zhao and Kun Zhan and Peng Jia and Xianpeng Lang and Hang Zhao},
booktitle={8th Annual Conference on Robot Learning},
year={2024},
url={https://openreview.net/forum?id=928V4Umlys}
}

@inproceedings{Sima2024ECCV,
  author    = {Chonghao Sima and Katrin Renz and Kashyap Chitta and Li Chen and Hanxue Zhang and Chengen Xie and Ping Luo and Andreas Geiger and Hongyang Li},
  title     = {DriveLM: Driving with Graph Visual Question Answering},
  booktitle = {European Conference on Computer Vision (ECCV)},
  year      = {2024}
}

@InProceedings{Hua_2025_CVPR,
  author    = {Hua, Hang and Liu, Qing and Zhang, Lingzhi and Shi, Jing and Kim, Soo Ye and Zhang, Zhifei and Wang, Yilin and Zhang, Jianming and Lin, Zhe and Luo, Jiebo},
  title     = {FINECAPTION: Compositional Image Captioning Focusing on Wherever You Want at Any Granularity},
  booktitle = {Proceedings of the IEEE/CVF Conference on Computer Vision and Pattern Recognition (CVPR)},
  year      = {2025},
  pages     = {24763-24773}
}

@article{brown2020language,
  title={Language models are few-shot learners},
  author={Brown, Tom and Mann, Benjamin and Ryder, Nick and Subbiah, Melanie and Kaplan, Jared D and Dhariwal, Prafulla and Neelakantan, Arvind and Shyam, Pranav and Sastry, Girish and Askell, Amanda and others},
  journal={Advances in neural information processing systems},
  volume={33},
  pages={1877--1901},
  year={2020}
}

@inproceedings{Wei2022CoT,
  title        = {Chain-of-Thought Prompting Elicits Reasoning in Large Language Models},
  author       = {Wei, Jason and Wang, Xuezhi and Schuurmans, Dale and Bosma, Maarten and Ichter, Brian and Xia, Fei and Chi, Ed and Le, Quoc and Zhou, Denny},
 booktitle = {Advances in Neural Information Processing Systems},
 editor = {S. Koyejo and S. Mohamed and A. Agarwal and D. Belgrave and K. Cho and A. Oh},
 pages = {24824--24837},
 publisher = {Curran Associates, Inc.},
 url = {https://proceedings.neurips.cc/paper_files/paper/2022/file/9d5609613524ecf4f15af0f7b31abca4-Paper-Conference.pdf},
 volume = {35},
 year = {2022}
}

@inproceedings{Kojima2022ZeroShotReasoners,
  title        = {Large Language Models are Zero-Shot Reasoners},
  author       = {Kojima, Takeshi and Gu, Shixiang Shane and Reid, Machel and Matsuo, Yutaka and Iwasawa, Yusuke},
 booktitle = {Advances in Neural Information Processing Systems},
 editor = {S. Koyejo and S. Mohamed and A. Agarwal and D. Belgrave and K. Cho and A. Oh},
 pages = {22199--22213},
 publisher = {Curran Associates, Inc.},
 url = {https://proceedings.neurips.cc/paper_files/paper/2022/file/8bb0d291acd4acf06ef112099c16f326-Paper-Conference.pdf},
 volume = {35},
 year = {2022}
}

@InProceedings{Johnson_2017_CVPR,
  author = {Johnson, Justin and Hariharan, Bharath and van der Maaten, Laurens and Fei-Fei, Li and Lawrence Zitnick, C. and Girshick, Ross},
  title = {CLEVR: A Diagnostic Dataset for Compositional Language and Elementary Visual Reasoning},
  booktitle = {Proceedings of the IEEE Conference on Computer Vision and Pattern Recognition (CVPR)},
  month = {July},
  year = {2017}
}

@article{kim2020interpretation,
  title={Interpretation of NLP models through input marginalization},
  author={Kim, Siwon and Yi, Jihun and Kim, Eunji and Yoon, Sungroh},
  journal={arXiv preprint arXiv:2010.13984},
  year={2020}
}

@inproceedings{xu2025llava,
  title={Llava-cot: Let vision language models reason step-by-step},
  author={Xu, Guowei and Jin, Peng and Wu, Ziang and Li, Hao and Song, Yibing and Sun, Lichao and Yuan, Li},
  booktitle={Proceedings of the IEEE/CVF International Conference on Computer Vision},
  pages={2087--2098},
  year={2025}
}

@inproceedings{li2023super,
  title={Super-clevr: A virtual benchmark to diagnose domain robustness in visual reasoning},
  author={Li, Zhuowan and Wang, Xingrui and Stengel-Eskin, Elias and Kortylewski, Adam and Ma, Wufei and Van Durme, Benjamin and Yuille, Alan L},
  booktitle={Proceedings of the IEEE/CVF conference on computer vision and pattern recognition},
  pages={14963--14973},
  year={2023}
}

@inproceedings{jain2019attention,
  title={Attention is not explanation},
  author={Jain, Sarthak and Wallace, Byron C},
  booktitle={Proceedings of the 2019 Conference of the North American Chapter of the Association for Computational Linguistics: Human Language Technologies, Volume 1 (Long and Short Papers)},
  pages={3543--3556},
  year={2019}
}

@article{sinha2025chart,
  title={Chart-RVR: Reinforcement Learning with Verifiable Rewards for Explainable Chart Reasoning},
  author={Sinha, Sanchit and Frunza, Oana and Rasul, Kashif and Nevmyvaka, Yuriy and Zhang, Aidong},
  journal={arXiv preprint arXiv:2510.10973},
  year={2025}
}

@inproceedings{sinha2025coco,
  title={COCO-Tree: Compositional Hierarchical Concept Trees for Enhanced Reasoning in Vision-Language Models},
  author={Sinha, Sanchit and Xiong, Guangzhi and Zhang, Aidong},
  booktitle={Proceedings of the 2025 Conference on Empirical Methods in Natural Language Processing},
  pages={2695--2711},
  year={2025}
}

\appendix

\clearpage
\newpage
\section{Appendix}
\label{sec:appendix}

\subsection{Dataset Details}
\label{app:dataset-details}

We evaluate on three datasets selected to span different forms of visually grounded reasoning.

\noindent\textbf{ChartQA.}
ChartQA \cite{masry2022chartqa} is a chart-based visual question answering benchmark where each example consists of a chart image (e.g., bar/line/pie charts) and a natural language question. Many questions require extracting precise numeric values, comparing trends, or aggregating quantities from the plotted visual marks and axes. This makes ChartQA particularly sensitive to failures of visual grounding, since language priors alone are often insufficient to answer correctly. We utilize 15000 random samples from the original train and 1500 from the test sets. To balance the impact of both trivial questions and complex multi-step queries, we ensure a balanced split of human (complex) and machine-labeled samples (trivial) as detailed in \cite{masry2022chartqa}.

\noindent\textbf{CLEVR.}
CLEVR \cite{Johnson_2017_CVPR} is a synthetic visual reasoning dataset composed of rendered scenes with simple objects and programmatically generated questions. Questions typically require compositional and multi-step reasoning (e.g., attribute comparisons, counting, spatial relations) with minimal linguistic ambiguity. Because the questions are controlled and the images contain the full evidence, CLEVR serves as a clean testbed for measuring whether CoT reasoning remains anchored to visual content. We utilize the scaled-down train and test sets consisting of 10000 train samples and 2000 test samples as sampled from the full dataset\footnote{\url{https://huggingface.co/datasets/dpdl-benchmark/clevr}}.

\noindent\textbf{CV-Bench.}
CV-Bench \cite{tong2024cambrian} is a multi-domain vision benchmark designed to probe general-purpose visual understanding and reasoning. It includes diverse image types and question styles that require a range of capabilities (e.g., recognizing visual entities, reading fine-grained visual details, and performing simple reasoning over them). Compared to ChartQA and CLEVR, CV-Bench is more heterogeneous and thus provides a stress test for whether improvements in CoT grounding and commitment dynamics generalize beyond a single visual domain. As the dataset only consists of a test set, we randomly split the dataset as available \footnote{\url{https://huggingface.co/datasets/nyu-visionx/CV-Bench}} into 1500 samples for training and 1140 for testing.

\subsection{Model Details}
\label{app:model-details}
\paragraph{Qwen2.5-VL (QVL-3B / QVL-7B).}
Both Qwen2.5-VL variants use a decoder-only Transformer LLM with grouped-query attention (GQA) and a ViT-style vision encoder whose patch tokens are projected into the LLM hidden space.
\emph{LLM (text) backbone:}
QVL-3B uses $L{=}36$ layers, hidden size $d{=}2048$, $h{=}16$ attention heads with $h_{kv}{=}2$ KV heads (GQA), and MLP intermediate size $d_{\text{ff}}{=}11008$; it supports up to 128k positions (\texttt{max\_position\_embeddings}=128000) and uses RoPE with large $\theta$ (\texttt{rope\_theta}=1e6).
QVL-7B uses $L{=}28$, $d{=}3584$, $h{=}28$, $h_{kv}{=}4$, and $d_{\text{ff}}{=}18944$, with the same 128k maximum positions and \texttt{rope\_theta}=1e6. 
\emph{Vision encoder:}
Both use a vision Transformer with depth 32 and width 1280, with 16 attention heads and intermediate size 3420, patch size 14, and local-window attention (\texttt{window\_size}=112) with periodic full-attention blocks at indices \{7,15,23,31\}. The vision features are mapped to the LLM hidden size via \texttt{out\_hidden\_size} (2048 for QVL-3B; 3584 for QVL-7B) and spatially merged with \texttt{spatial\_merge\_size}=2. 

\paragraph{InternVL3.5 (IVL-4B / IVL-8B).}
InternVL3.5 is released as an \texttt{internvl\_chat} model with (i) an LLM backbone (here, Qwen3) and (ii) an InternViT vision encoder.
\emph{LLM (Qwen3) backbone:}
IVL-4B uses $L{=}36$, hidden size $d{=}2560$, $h{=}32$ attention heads with $h_{kv}{=}8$ KV heads, head dimension 128, and MLP intermediate size $d_{\text{ff}}{=}9728$; it supports up to 40960 positions.
IVL-8B uses $L{=}36$, $d{=}4096$, $h{=}32$, $h_{kv}{=}8$, head dimension 128, and $d_{\text{ff}}{=}12288$, with the same 40960 maximum positions. 
\emph{Vision encoder:}
Both use \texttt{InternVisionModel} (InternViT-6B-style) with $L_v{=}24$ layers, hidden size 1024, 16 attention heads, and MLP intermediate size 4096, operating at image size 448 with patch size 14. 
\emph{Image tokenization / dynamic patches:}
The configs enable dynamic image sizing (\texttt{dynamic\_image\_size}=true) with \texttt{force\_image\_size}=448, \texttt{min\_dynamic\_patch}=1 and \texttt{max\_dynamic\_patch}=12, and \texttt{downsample\_ratio}=0.5.

\paragraph{Gemma3 (Gemma-4B / Gemma-12B).}
Gemma3 is released as a multimodal \texttt{Gemma3ForConditionalGeneration} model with (i) a Gemma3 text decoder and (ii) a SigLIP-style vision encoder. 
\emph{LLM backbone:}
Gemma-4B uses $L{=}34$, hidden size $d{=}2560$, $h{=}8$ attention heads with $h_{kv}{=}4$ KV heads, head dimension 256, and MLP intermediate size $d_{\text{ff}}{=}10240$.
Gemma-12B uses $L{=}48$, hidden size $d{=}3840$, $h{=}16$ attention heads with $h_{kv}{=}8$ KV heads, head dimension 256, and MLP intermediate size $d_{\text{ff}}{=}15360$.
Both support up to 131072 positions (128K context), use RoPE scaling with factor 8.0, and interleave local sliding-window attention with \texttt{sliding\_window}=1024.
\emph{Vision encoder:}
Both Gemma-4B and Gemma-12B use a \texttt{SiglipVisionModel}-style encoder with $L_v{=}27$ layers, hidden size 1152, 16 attention heads, and MLP intermediate size 4304, operating at image size 896 with patch size 14.
\emph{Image tokenization / multimodal interface:}
Gemma3 uses \texttt{mm\_tokens\_per\_image}=256, with dedicated begin/end/image token indices, and processes images at a fixed 896$\times$896 resolution; optional pan-and-scan inference further crops high-resolution or non-square images into multiple smaller patches before concatenating them with the base image embedding.

\noindent \textbf{Loss Coefficient Weights.} We set the coefficients of all loss weights to 1.0 as the weights are self-normalized.

\subsection{Chain-of-Thought Data Construction}
\noindent \textbf{Data Annotation:} To construct the ground-truth CoT supervision, we use a fixed system prompt template as shown in Figure~\ref{fig:dset-gen-prompt} to elicit step-by-step rationales from OpenAI API for `gpt-5-mini-2025-08-07'. For each training instance, we provide the image, the query, and the gold answer, and instruct the model to generate a concise sequence of atomic steps labeled step-1 through step-n. The template enforces a consistent format and encourages each step to perform a single, simple operation (e.g., locating an object, checking an attribute, comparing values), yielding structured rationales that are easy to parse and align with token-level analyses during fine-tuning.

\noindent \textbf{Multi-agent Ensemble Check:} To improve the quality of the annotated CoT traces, we employ a two-agent pipeline with \texttt{gpt-5-mini-2025-08-07}. The first agent generates an initial step-by-step rationale from the image, query, and gold answer, while the second agent acts as an oracle verifier that inspects the generated trace for logical consistency, visual faithfulness, and answer correctness. If the trace contains unsupported, incomplete, or inconsistent reasoning, the verifier revises it accordingly; otherwise, it is accepted unchanged. This verification stage provides an additional layer of automatic quality control before manual inspection, reducing noise in the final CoT supervision.

\noindent \textbf{Manual Check for consistency:} To ensure the accuracy of generated CoT data, we also conduct a manual 2-pass check detailed below:
\begin{itemize}
    \item For pass 1, all samples with overtly short/long and incorrect reasoning traces are filtered (i.e. trace length less than 3 or more than 8 based on manual inspection and outlier rejection). To validate if a trace is wrong, we algorithmically check if the last CoT line trace contains the correct answer.
    \item For pass 2, two doctoral-level researchers are employed to validate the reasoning traces of a randomly sampled data subset (1000 samples). Each researcher filtered fewer than 6 samples (1\%) in total, with a 100\% agreement between them.
\end{itemize}

\begin{figure}[h]
\centering
\begin{tcolorbox}
[colframe=black,title=System Prompt template for CoT Dataset Generation]
You are a vision-language assistant. You will be given an image, query and answer.  \\
You have to output how we can answer the query step-by-step and arrive at the answer. \\ 
Each step should be labeled as "step-1", ... "step-n". 
Keep each step simple and concise and only doing one thing at a time. \\ 
Input Format: \\
Query: query \\ 
Answer: answer \\
Output Format: \\
step-1:  \\
step-2: \\
... \\
step-n: \\
\end{tcolorbox}
\caption{\textbf{System prompt template} for generating the Ground Truth CoT Data}
\label{fig:dset-gen-prompt}
\vspace{-10pt}
\end{figure}

\subsection{Inference Procedure}

\subsubsection{Structured System Prompt}
To enforce a consistent and standard output format, we utilize the same system prompt shown in Figure~\ref{fig:inference-prompt}. This ensures that the free-form CoT, which can differ wildly amongst model families, has a fixed algorithmic reasoning procedure.

\begin{figure}[h]
\centering
\begin{tcolorbox}
[colframe=black,title=System Prompt template for CoT Inference]
You are a vision-language assistant. You are given an image and a query.
    Think step by step and then try to answer the question with short words or phrases if possible.
    Output Format: \\
    $<$think$>$ \\
    step-1: \\
    step-2: \\
    ... \\
    $<$/think$>$ \\
    $<$answer$>$\\
    Answer the question\\
    $<$/answer$>$
\end{tcolorbox}
\caption{\textbf{System prompt template} for Chain-of-Thought inference.}
\label{fig:inference-prompt}
\vspace{-10pt}
\end{figure}

\subsubsection{Inference Settings}
\noindent \textbf{Attention Implementation:} For all our experiments, we utilize `eager' attention implementation, which, albeit slower, gives a deterministic value to all computed attention weights as opposed to `FlashAttention'. 

\noindent \textbf{Decoding:} For QVL family, we utilize a maximum number of generated tokens of 256 for CoT and 32 for direct across all datasets. As IVL family usually has a more verbose style of native CoT, we utilize a maximum of 512 tokens for CoT across all datasets. We utilize a temperature value set to 0 for all experiments to ensure deterministic outputs.

\noindent \textbf{Image Size:} To enforce standard layout, we enforce all images to be between 360 * 28 * 28 and 480 * 28 * 28 pixels, which corresponds to about (512,512) image sizes.

\noindent \textbf{Accuracy:} For ChartQA, we utilize the Relaxed Accuracy metric as proposed in \cite{masry2022chartqa}. Relaxed Accuracy considers the predicted numerical answers within a tight threshold as correct (not an exact match). The threshold value is set as 0.05 (5\%). Mathematically for a prediction $\hat{y}$ and ground truth $\hat{y*}$, 
\[\texttt{Relaxed Accuracy} (\hat{y},{y*}) =  \mathbf{1}[\tfrac{|y*-\hat{y}|}{y*} \leq 0.05], \]
where $\mathbf{1}$ is the Indicator Function. Note that an exact match is utilized for non-numeric or mixed alphanumeric answers. For all other datasets, we utilize RA with tolerance 0.0.

\subsubsection{Output Span Description}
\label{app:span-defs}
We decompose each model generation into two contiguous token spans: a \textbf{think span} and an \textbf{answer span} enforced by system prompt in Figure~\ref{fig:inference-prompt}. The \textbf{think span} consists of all tokens emitted as the chain-of-thought - i.e., the intermediate reasoning tokens generated \emph{before} the model begins producing the final answer. The \textbf{answer span} consists of the final answer tokens only, i.e., the tokens that realize the predicted answer string. Concretely, we enforce a fixed delimiter in the decoding template \texttt{</think>} and define the \textbf{think span} as all generated tokens from the first generated token up to (but excluding) the delimiter, and the \textbf{answer span} as the tokens generated after the delimiter until the end-of-answer (EOS) token. When the model produces multi-token answers, the answer span includes the entire answer token sequence. For direct-answer baselines (no-CoT), the think span is empty and the full generation is treated as the answer span.

\subsection{CoT Performance across model scales}
\begin{figure*}[h]
\begin{minipage}[t]{0.39\textwidth}
  \centering
  \includegraphics[width=0.95\linewidth]{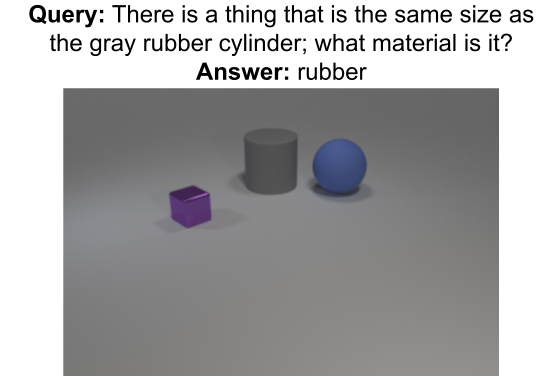}
\end{minipage}\hfill
\begin{minipage}[t]{0.61\textwidth}
  \centering
  \includegraphics[width=\linewidth]{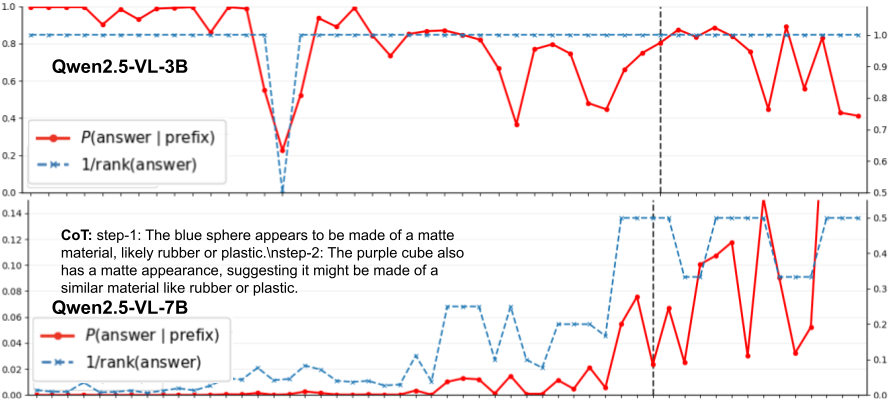}
\end{minipage}
\caption{\textbf{Early answer commitment differs by scale.}  Qwen2.5-VL-3B exhibits high confidence early in the think span (premature commitment), whereas Qwen2.5-VL-7B delays commitment until later. The dashed vertical line marks the transition from the think span to the answer span.}
\label{fig:base-answer-commit-clevr}
\vspace{-10pt}
\end{figure*}

Figure~\ref{fig:base-answer-commit-clevr} shows an example from CLEVR. In this sample, the SFT model exhibits \emph{premature answer commitment} during the thinking span. In contrast, Att-CoT suppresses these early commitment spikes and concentrates confidence closer to the answer span, yielding a correct response with a more coherent, stepwise rationale.

\subsection{Diverse Robust Baselines without Zero image for Vis-Dep}
To complement our attention-based grounding analysis, we evaluate model robustness under a \textbf{Gaussian blur} transformation applied to the input image. Gaussian blur suppresses high-frequency visual details such as fine edges, small text, and local texture cues, while largely preserving the global scene layout and coarse object structure. Intuitively, if a model genuinely relies on precise visual evidence during reasoning, its performance should degrade under this perturbation, especially on visually intensive tasks that require fine-grained perception. We therefore use Gaussian blur as a simple but effective stress test for visual dependence, allowing us to assess whether different training strategies encourage stronger reliance on detailed image content rather than superficial language priors.

In Table~\ref{tab:gaussian-blur-results}, we report the Vis-Dep scores under Gaussian blur for QVL-7B, IVL-8B, and Gemma-12B on ChartQA and CLEVR datasets. Across all three backbones, Att-CoT consistently achieves the highest Vis-Dep among trained methods on both ChartQA and CLEVR, indicating stronger reliance on fine-grained visual evidence even under degraded image quality. In contrast, SFT, LLaVA-CoT, and LLaVA-R generally exhibit substantially lower scores, suggesting a greater tendency to fall back on textual priors when detailed visual cues are suppressed. The trends follow similar insights from the zero-image experiments, implying our Vis-Dep metric is robust to baseline no-information image.

\begin{table*}[h]
\centering
\resizebox{0.75\textwidth}{!}{
\begin{tabular}{l|cc|cc|cc}
\toprule
\multirow{2}{*}{\textbf{Method}} 
& \multicolumn{2}{c|}{\textbf{QVL-7B}} 
& \multicolumn{2}{c|}{\textbf{IVL-8B}} 
& \multicolumn{2}{c}{\textbf{Gemma-12B}} \\
& \textbf{ChartQA} & \textbf{CLEVR}
& \textbf{ChartQA} & \textbf{CLEVR}
& \textbf{ChartQA} & \textbf{CLEVR} \\
\midrule
CoT (no training) & 0.93  & 0.55 & 0.79 & 0.61  & 0.80 & 0.54 \\
\midrule
SFT         & 0.54 & 0.34 & 0.49 & 0.07  & 0.68 & 0.19  \\
LLaVA-CoT   & 0.48  & 0.38 & 0.50 & 0.11  & 0.67 & 0.08 \\
LLaVA-R  & 0.50 & 0.31 & 0.44 & 0.12  & 0.61 & 0.15 \\
\textbf{Att-CoT} & \bf 0.78 & \bf 0.48 & \bf 0.64  & \bf 0.21 & \bf 0.74 & \bf 0.38 \\
\bottomrule
\end{tabular}
}
\caption{Performance under Gaussian blur across different MLLM backbones and training methods on ChartQA and CLEVR.}
\label{tab:gaussian-blur-results}
\end{table*}

\subsection{Baseline Replication Details}

\noindent \textbf{LLaVA-COT \cite{xu2025llava}:} LLaVA-COT was originally developed on top of earlier-generation LLaVA backbones which had extremely low CoT reasoning abilities. Since these models predate more recent multimodal architectures with stronger native reasoning and instruction-following capabilities, we replicate the method for our experiments on three modern MLLM families: Qwen2.5-VL, InternVL3.5, and Gemma3. These backbones are substantially more CoT-friendly out of the box, making them a stronger and more relevant testbed for evaluating whether the gains of the original approach persist under current-generation multimodal models. We format the data as described by prompting API to generate \texttt{<SUMMARY></SUMMARY>,
<CAPTION></CAPTION>, <REASONING></REASONING>} \\ and \texttt{
<CONCLUSION></CONCLUSION>} tags. For test time scaling using `Stage-wise retracing search', we use $N=4$, i.e., Best-of-4. Interestingly, the improvements over SFT are meager - implying modern backbones do not benefit as much from test-time scaling as LLaVA models.

\noindent \textbf{LLaVA-R \cite{zhang2025improve}: } LLaVA-R-DPO was also introduced on older LLaVA-family backbones. We incorporate the training setup of `LLAVA-REASONER-SFT' verbatim from the GitHub repo\footnote{\url{https://github.com/RifleZhang/LLaVA-Reasoner-DPO}}. We re-use the DPO data for ChartQA. For CLEVR and CV-Bench, we compare against the SFT approach.

\subsection{Detailed Computational Cost Analysis}
Let the decoder comprise $L$ transformer layers with $H$ attention heads per layer, hidden size $d$, visual-token index set $\mathcal{V}$ with cardinality $|\mathcal{V}| = N_v$, and total sequence length $S = T + K$, where $T$ denotes the number of CoT tokens and $K$ the number of answer tokens. Standard CoT-SFT optimizes only the autoregressive negative log-likelihood objective $\mathcal{L}_{\mathrm{SFT}}$, whose dominant cost is the usual forward-backward pass through the MLLM. In contrast, our method augments training with two auxiliary terms,
\[
\mathcal{L}_{\mathrm{Att\text{-}CoT}}
=
\mathcal{L}_{\mathrm{SFT}}
+
\mathcal{L}_{\mathrm{VG}}
+
\mathcal{L}_{\mathrm{DC}},
\]
while introducing no additional trainable parameters, no extra model branches, and no second decoding pass. The visual-grounding term reuses the native attention maps $A^{(l,h)}_{t,j}$ already produced during decoding, and computes the average visual attention mass at rationale step $t$ as
\[
\alpha_t^{\mathrm{vis}}
=
\frac{1}{LH}
\sum_{l=1}^{L}
\sum_{h=1}^{H}
\sum_{j \in \mathcal{V}}
A^{(l,h)}_{t,j},
\]
which adds only a tensor-reduction overhead of order $\mathcal{O}(TLHN_v)$. Similarly, the delayed-commitment term operates on the gold-answer confidence
\[
\pi_t = p_{\theta}(y^\star \mid I,q,r_{<t}),
\]
obtained directly from the model logits, and therefore incurs only $\mathcal{O}(T)$ additional elementwise operations across CoT positions. Consequently, the asymptotic training complexity remains identical to standard SFT, with the extra cost dominated by lightweight reductions over tensors that are already materialized in memory. 
Hence, if the baseline SFT cost is denoted by $\mathcal{C}_{\mathrm{SFT}}$, the total cost of Att-CoT can be written as
\[
\mathcal{C}_{\mathrm{Att\text{-}CoT}}
=
\mathcal{C}_{\mathrm{SFT}}
+
\mathcal{O}(TLHN_v),
\]
which preserves the same dominant asymptotic scaling as vanilla CoT fine-tuning. In practice, $\mathcal{O}(TLHN_v) << \mathcal{C}_{\mathrm{SFT}}$.

\subsection{Difficulty aware Splits}
In Table~\ref{tab:difficulty-aware}, we report the performance on the Human (complex) and Machine (trivial) splits of the ChartQA dataset. Our sampled test-set for ChartQA consists of 750 samples each from Human and Machine subsets. Across three backbones, Att-CoT achieves the best performance on both splits, with the gains being consistently larger on the Human subset than on the Machine subset i.e. our method helps on queries that require deeper reasoning. For example, on QVL-3B, Att-CoT improves over SFT by only $+0.2$ on Machine questions but by $+4.3$ on Human questions; similarly, the gain on Gemma-4B is $+0.8$ on Machine versus $+3.3$ on Human.
\begin{table*}[h]
\centering
\resizebox{0.55\textwidth}{!}{
\begin{tabular}{l|cc|cc|cc}
\toprule
\centering
\multirow{2}{*}{\textbf{Method}} 
& \multicolumn{2}{c|}{\textbf{QVL-3B}} 
& \multicolumn{2}{c|}{\textbf{IVL-4B}} 
& \multicolumn{2}{c}{\textbf{Gemma-4B}} \\
\centering
& \textbf{M} & \textbf{H}
& \textbf{M} & \textbf{H}
& \textbf{M} & \textbf{H} \\
\midrule
SFT         & 92.0 & 70.4 & 95.2 & 77.5 & 91.4 & 69.5  \\
LLaVA-CoT   & 91.2 & 65.4 &  94.9 & 74.7 & 91.2 & 66.0 \\
LLaVA-R     & 91.7 & 68.6 & 94.9 & 76.3 & 91.3 & 67.4  \\
\textbf{Att-CoT}     & \bf 92.2 & \bf 74.7 & \bf 95.3 & \bf 78.3 & \bf 92.2 & \bf 72.8    \\
\bottomrule
\end{tabular}
}
\caption{Comparison of different training methods across multiple MLLM backbones under machine and human sub-splits of ChartQA.}
\label{tab:difficulty-aware}
\end{table*}

\subsection{Hyperparameter Setup}
\noindent \textbf{Training Configuration}
For both SFT and Att-CoT, we train the model for a maximum of 5 epochs and select the model with the lowest loss on the validation set. For comparison baselines, we utilize the hyperparameters mentioned in the respective method implementation descriptions. Additionally, we utilize the AdamW optimizer for training with a training batch size of 4 across 2 NVIDIA A100 GPUs. The learning rate schedule utilized is linear with a maximum learning rate of $1e-5$ and 1000 warmup steps. During training, the entire model is trained with FP16 precision, and the vision tower is trained using an LR of $2e-6$. As is standard practice, all image-relevant tokens are masked during fine-tuning. The prompt to elicit CoT response is the same prompt shown in Figure~\ref{fig:inference-prompt}.

\subsection{Answer Commitment Visual Analysis}
\begin{figure*}[h]
    \centering
    \includegraphics[width=0.7\textwidth]{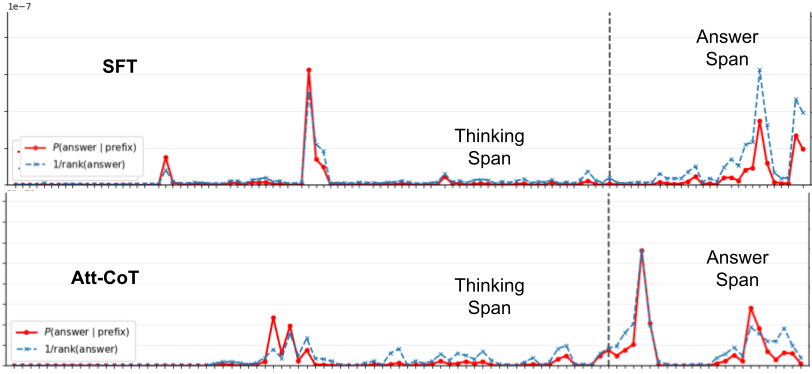}
    \caption{Representative example where SFT exhibits intermittent spikes during the thinking span, indicating premature commitment to the final answer, whereas Att-CoT suppresses early peaks and concentrates commitment closer to the answer span.}
    \label{fig:chartqa-sft-att-cot}
\end{figure*}

Figure~\ref{fig:chartqa-sft-att-cot} visualizes answer commitment for SFT vs Att-CoT, illustrating how Att-CoT delays answer commitment until the end of reasoning. Figure~\ref{fig:base-att-cot-heatmap} visualizes attention mass on visual tokens across layers for SFT vs Att-CoT, illustrating how Att-CoT encourages more sustained visual token access during the think span.

\subsection{Visual Attention Analysis across MLLM scales}
\begin{figure*}[h]
    \centering
    \includegraphics[width=0.8\textwidth]{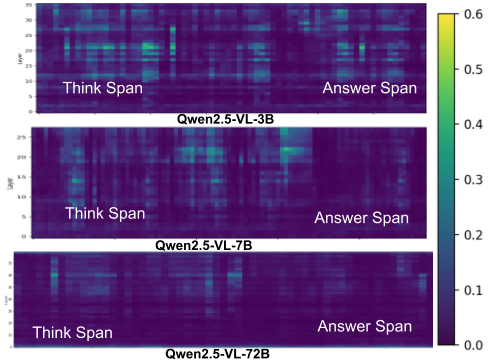}
    \caption{\textbf{Visual Token Attention across model scales.} We plot layer-step heatmaps of the average attention mass assigned to visual tokens, $A_{t,\mathrm{vis}}$, for Qwen2.5-VL-\{3B, 7B, 72B\} on the same CoT-prompted example. The horizontal axis is the generation step $t$, the vertical axis is the decoder layer, and color indicates $A_{t,\mathrm{vis}}$ (higher = more cumulative visual attention). The \emph{think} and \emph{answer} spans are delineated to show how visual grounding evolves during think and answer generation.}
    \label{fig:vta-scales}
\end{figure*}

Next, Figure~\ref{fig:vta-scales} reports VTA heatmaps across model scales, showing how visual attention mass varies across layers and generation steps. Across Qwen2.5-VL (3B, 7B, 72B) on the same CoT-prompted example, smaller models display more sporadic and uneven visual grounding across steps, while larger models exhibit more structured, distributed attention patterns across layers, with grounding that is generally more stable across the trajectory. 

\subsection{How Att-CoT changes visual attention as compared to base model}
\begin{figure*}[h]
    \centering
    \includegraphics[width=0.75\linewidth]{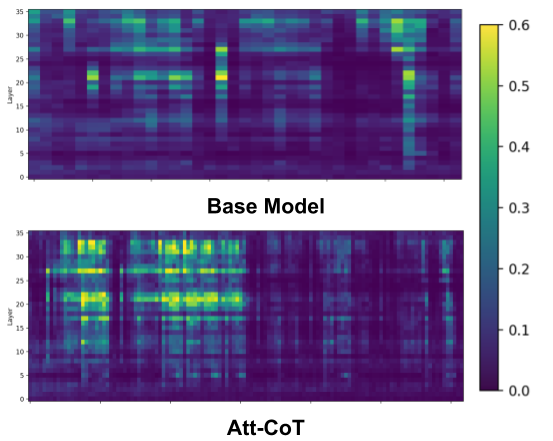}
    \caption{\textbf{Attentive-CoT increases sustained visual attention during reasoning.} Layer-step heatmaps of $A_{t,\mathrm{vis}}$ for the base model (top) versus Att-CoT (bottom) on the same CoT-prompted sample. Att-CoT shifts attention mass toward the think span and maintains more consistent visual grounding across steps, compared to the base model's more transient/spiky visual attention near the answer span.}
    \label{fig:base-att-cot-heatmap}
\end{figure*}

In Figure~\ref{fig:base-att-cot-heatmap}, we demonstrate the change in visual attention mass of Att-CoT as compared to the base model. We observe that Att-CoT has higher attention weights on the \texttt{think} span. This observation, along with higher accuracies and improved qualitative reasoning performance, Att-CoT is an effective tool to improve fine-tuning performance.

\subsection{Qualitative Examples}
\begin{figure*}[h]
    \centering
    \includegraphics[width=\textwidth]{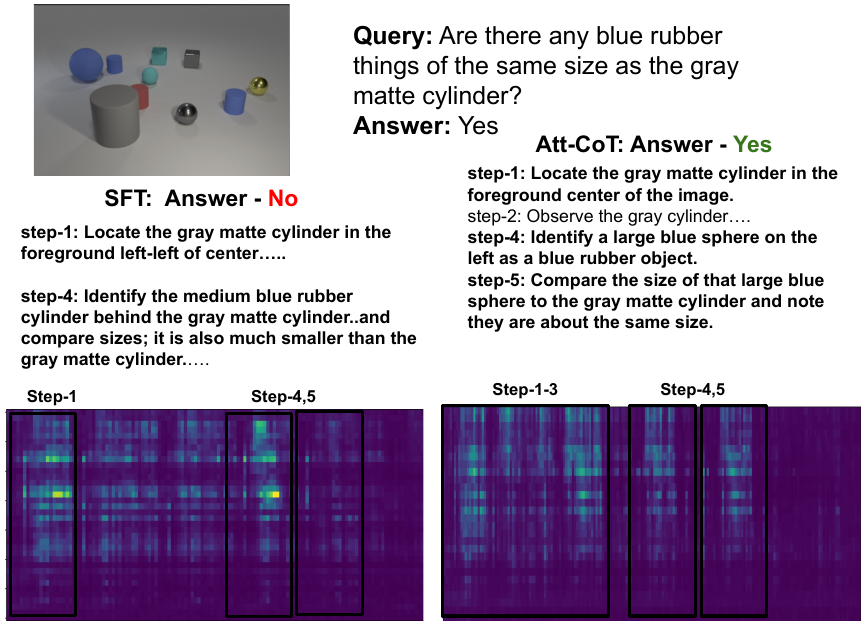}
    \caption{Qualitative comparison of SFT vs. Att-CoT on a CLEVR example. SFT model commits to an incorrect answer (No) and produces a loosely grounded rationale, whereas Att-CoT correctly answers (Yes) with a more faithful size-comparison trace. The bottom heatmaps (token-to-visual attention over the CoT trajectory, aggregated across heads) show that Att-CoT sustains stronger, more structured visual grounding during reasoning, while SFT has multiple spikes in between sparse values and also spikes during answer. Black boxes highlight high-attention regions over visual tokens.}
    \label{fig:qualitative-examples}
\end{figure*}

\begin{figure*}[h]
    \centering
    \includegraphics[width=\textwidth]{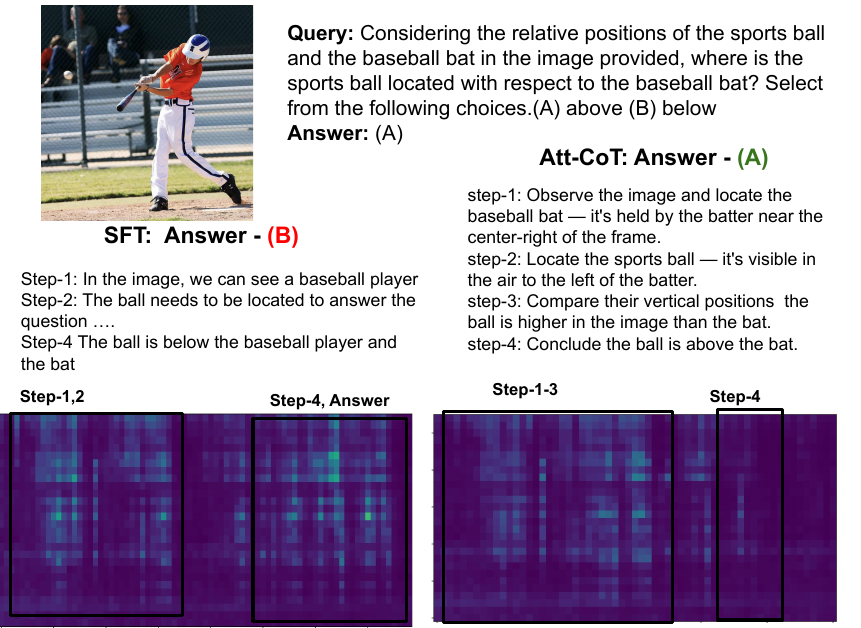}
    \caption{Qualitative comparison of SFT vs. Att-CoT on a CV-Bench example. SFT model commits to an incorrect answer (Below) and produces a loosely grounded rationale, whereas Att-CoT correctly answers (Above) with a more faithful stepwise spatial comparison. The bottom heatmaps (token-to-visual attention over the CoT trajectory, aggregated across heads) show that Att-CoT sustains stronger, more structured visual grounding during reasoning, especially when evaluating the locations of the objects in the image, while SFT focuses attention mass near the answer span. Black boxes highlight high-attention regions over visual tokens.}
    \label{fig:qualitative-examples-2}
\end{figure*}

Figures~\ref{fig:qualitative-examples} and \ref{fig:qualitative-examples-2} show qualitative comparisons between SFT and Att-CoT. We visualize the generated spans and the corresponding attention behavior to highlight cases where Att-CoT improves both correctness and grounding. Cumulative attention mass over visual tokens improves accurate spatial understanding - making Att-CoT outperform SFT. 

\subsection{Attention-based Grounding Analysis}
\begin{figure*}
    \centering
    \includegraphics[width=\textwidth]{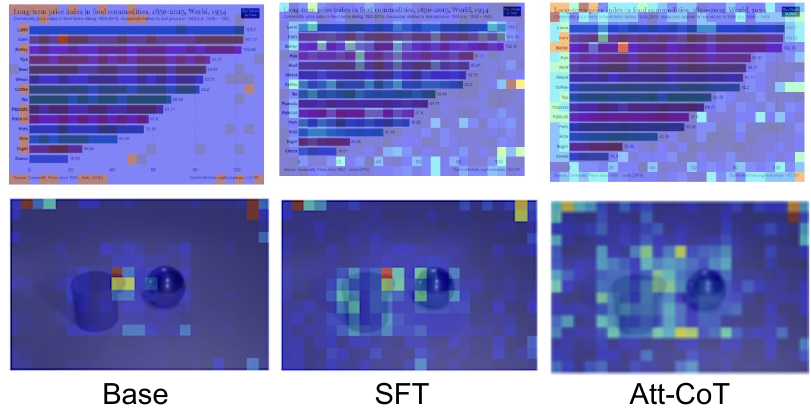}
    \caption{Comparison of token-to-visual attention overlays for the base model, SFT, and Att-CoT on the same example. (TOP) Query: `Number of items in the chart?' (BOTTOM) 'Where is the cylinder with respect to the shiny sphere?'. We visualize the cumulative visual attention induced by the generated CoT trajectory, together with zoomed-in views of the relevant region. Relative to the base model and SFT, Att-CoT produces more spatially concentrated and semantically aligned attention on the task-relevant visual evidence, indicating stronger grounding during reasoning rather than only near the final answer span.}
\label{fig:attn-overlay}
\end{figure*}

Figure~\ref{fig:attn-overlay} compares the spatial distribution of token-to-visual attention for the base model, SFT, and Att-CoT on the same sample from ChartQA (TOP) and CLEVR (BOTTOM). We observe that Att-CoT assigns more concentrated attention mass to the visually relevant region, whereas the base model and SFT exhibit more diffuse or weakly localized patterns. This qualitative trend is consistent with our broader visual-attention analysis, where Att-CoT increases attention over the \textit{think} span and improves grounding during multi-step reasoning. Together, these overlays suggest that Att-CoT encourages more faithful and task-aligned visual attention than standard fine-tuning.


\end{document}